%% file: main.tex
\documentclass[10pt,twocolumn,letterpaper]{article}

\usepackage{iccv_style/iccv}      
\usepackage[accsupp]{axessibility}  

\input{math_commands.tex}


\PassOptionsToPackage{numbers, compress, }{natbib}

\usepackage[linesnumbered,ruled]{algorithm2e}
\SetAlFnt{\small}
\SetAlCapFnt{\small}
\SetAlCapNameFnt{\small}
\usepackage{algorithmic}
\algsetup{linenosize=\tiny}

\makeatletter
\usepackage{tcolorbox}

\usepackage{amssymb}
\usepackage{utfsym}

\makeatother

\usepackage[utf8]{inputenc} 
\usepackage[T1]{fontenc}    
\usepackage{url}            
\usepackage{booktabs}       
\usepackage{amsfonts}       
\usepackage{nicefrac}       
\usepackage{microtype}      
\usepackage{xcolor}         
\usepackage{graphicx}
\usepackage{graphics}
\usepackage{xcolor}
\usepackage{subcaption}
\usepackage{amsmath}
\usepackage{multirow}
\usepackage{wrapfig}
\usepackage{enumitem}
\usepackage{float}
\definecolor{cerisepink}{rgb}{0.93, 0.23, 0.51}

\definecolor{iccvblue}{rgb}{0.21,0.49,0.74}
\usepackage[pagebackref,breaklinks,colorlinks,allcolors=iccvblue]{hyperref}

\def\paperID{15092}
\def\confName{ICCV}
\def\confYear{2025}

\author{Yijun Liang\thanks{These authors contributed equally to this work.}, Shweta Bhardwaj\footnotemark[1], Tianyi Zhou\\
University of Maryland, College Park\\
{\tt\small {\{yliang17,shweta12\}}@umd.edu, tianyi.david.zhou@gmail.com}\\
\small{Project: \url{https://github.com/tianyi-lab/DisCL}}
}
\title{Diffusion Curriculum:\\Synthetic-to-Real Data Curriculum via Image-Guided Diffusion}

\begin{document}
\maketitle

\begin{abstract}
\label{sec:abstract}
Low-quality or scarce data has posed significant challenges for training deep neural networks in practice. While classical data augmentation cannot produce very different new data, diffusion models open up a new door to build self-evolving AI by generating high-quality and diverse synthetic data through text-guided prompts. However, text-only guidance cannot control synthetic images' proximity to the original images, resulting in out-of-distribution data detrimental to model performance. To overcome the limitation, we study image guidance to achieve a spectrum of interpolations between synthetic and real images. With stronger image guidance, the generated images are similar to the training data, but are hard to learn. With weaker image guidance, the synthetic images will be easier to learn but suffer from a larger distribution gap to the original data. The generated full spectrum of data enables us to build a novel ``\textbf{Di}ffu\textbf{s}ion \textbf{C}urricu\textbf{L}um (DisCL)''. DisCL adjusts the image guidance level of image synthesis for each training stage: It identifies and focuses on hard samples for the model and assesses the most effective guidance level of synthetic images to improve hard data learning. We apply DisCL to two challenging tasks: long-tail (LT) classification and learning from low-quality data. It focuses on lower-guidance images of high quality to learn prototypical features as a warm-up for learning higher-guidance images that might be weak on diversity or quality. DisCL achieves a gain of 2.7$\%$ and 2.1$\%$ in OOD and ID macro-accuracy when applied to iWildCam dataset. On ImageNet-LT, DisCL improves the base model's tail-class accuracy from 4.4$\%$ to 23.64$\%$ and leads to a 4.02$\%$ improvement in all-class accuracy.\looseness-1
\end{abstract}

\input{sections/introduction}

\input{sections/related_works}

\input{sections/sec_3_method}
\input{sections/experiments}
\input{sections/ablation_analysis}

\input{sections/conclusion}



\medskip

\bibliography{bibtext}
\bibliographystyle{iccv_style/ieeenat_fullname}
\newpage
\appendix

\vspace*{-0.1cm}
\maketitlesupplementary
\input{sections/appendix}

\end{document}

%% file: math_commands.tex
\usepackage{amsmath,amsfonts,bm}

\def\eqref#1{equation~\ref{#1}}

\def\1{\bm{1}}

\DeclareMathAlphabet{\mathsfit}{\encodingdefault}{\sfdefault}{m}{sl}
\SetMathAlphabet{\mathsfit}{bold}{\encodingdefault}{\sfdefault}{bx}{n}



%% file: sections/introduction.tex
\section{Introduction}\label{sec:introduction}
\vspace*{-0.2cm}
While existing machine learning approaches can train representation or discriminative models with promising generalization performance, their success highly relies on the quality and quantity of the training data. However, in enormous practical scenarios, the data are collected from real environments so neither the quality nor the quantity can always be guaranteed. For example, it is difficult to control the light conditions, weather, motion blur, or the position of objects in the scenes captured by trail/animal cameras, traffic cameras, motion cameras, or robot cameras. Likewise, it is also difficult to keep different classes in the collected data balanced so the model may perform much poorer on tail classes with scarce data. On the other hand, the low-quality/quantity of data also makes the model more prone to the gap between the test and training distributions, thereby posing an out-of-distribution challenge. 
In many cases, such ``hard'' training data hinders effective learning, introduces biases or outliers, and may even impact the learning of other data. \looseness-1

Data augmentation and synthesis have been studied to address the challenges of hard real data. 
By applying pre-defined transformations \citep{ahn2023cuda} to data in scarce classes 
or modifying their backgrounds \citep{beery2020synthetic,gao2022out}, data augmentation helps learn representations robust to these task-irrelevant variations. While the augmented data may lack sufficient diversity or non-trivial difference to the original data, the recent text-to-image generative models such as GAN or Stable Diffusion enable more sophisticated data synthesis \citep{dunlap2024diversify} of diverse higher-quality samples, while the text prompts retain the task-related features. 
For instance, recent works \citep{han2024latent, play-no-favorites-Um-Ye-2023, datadream2024} have focused on training specialized diffusion models specifically for sampling from underrepresented or hard classes to increase diversity.
However, these existing methods are still challenged when scaling to real-world data, such as in-the-wild or long-tail learning scenarios, where the data distribution is highly imbalanced, diverse, and unpredictable.
Although text-to-image synthesis improves the data quality and quantity, the synthetic data are solely controlled by text prompts but lack sufficient visual similarity to the original image, 
which leads to a distribution gap to the original data and hurts the generalization performance. \looseness-1
To maximize the merits of synthetic data for learning hard data in real applications and address the syn-to-real gap, we harness the image guidance in diffusion models to generate a full spectrum of interpolations between synthetic data (\textit{i.e.,} generated only from text prompts) and real data (\textit{i.e.,} original images that may suffer from low-quality or sparse quantity).
The synthetic data at each level of interpolation are generated under the weighted guidance of both the text prompt (e.g., the class name) and the real images.
While stronger image guidance preserves visual similarities to the original image, for low-quality or low-quantity data, weaker image guidance could lead to high-quality, diverse, and potentially easier (e.g., with prototypical features) data. 
Hence, the syn-to-real interpolations create a novel space of synthetic data to design a \textbf{generative curriculum} that can adjust the quality, diversity, and/or difficulty of data for different training stages, by selecting the guidance level according to a pre-defined schedule or training dynamics.

In this paper, we develop novel generative curriculum learning approaches for two types of challenging applications with ``hard'' real images: long-tail classification and learning from low-quality images.
In \textit{long-tail classification}, learning the tail classes' features is challenging due to their data deficiency and the lack of diversity compared to ``head classes''. To address this challenge, we propose a curriculum that first learns synthetic images with lower image guidance for tail classes since they enhance the diversity and quantity of the original data. The curriculum then gradually increases the guidance level and learns synthetic images closer to the original images, thereby progressively bridging the syn-to-real gap.
In \textit{learning from low-quality data}, the primary challenge is to capture the critical features of the target classes, which is hard due to intricate background, occlusion, or motion blur in the original images.
In contrast, images generated with lower image guidance usually contain prototypical features easier to learn.
That being said, an overly high or low guidance level may enlarge the domain gap between the training data and the target (in-distribution or out-of-distribution) data. To avoid negative transfer caused by the domain gap and to maximize the merits of synthetic data, we develop an adaptive curriculum that selects the guidance level of synthetic data leading to the greatest progress of each training stage.   

We examine two DisCL curricula on benchmark datasets,  WILD-iWildCam \citep{beery2021iwildcam} and ImageNet-LT \citep{liu2019large}, for learning from low-quality images, and long-tail classification respectively. Our DisCL curricula improve OOD and ID accuracy by 2.7\% and 2.1\% respectively on iWildCam. On ImageNet-LT, DisCL improves the minority classes' accuracy by 19.24\% and leads to a 4.02\% improvement in the overall accuracy.
Our main contributions can be summarized as follows: \looseness-1
\begin{itemize}[leftmargin=*]
    \item Harness image guidance in diffusion models to systematically create a spectrum of synthetic-to-real data for each sample, enabling the design of effective training curricula to address hard data learning. 
    
    \item Propose the ``\textbf{Di}ffu\textbf{s}ion \textbf{C}urricu\textbf{L}um (DisCL)'' paradigm, which selects synthetic data of different guidance levels to meet the needs of each training stage. We propose two novel DisCL curricula to address two key applications: long-tail classification and learning from low-quality data. 
    
    \item Examine the two DisCL curricula on challenging datasets and demonstrate their effectiveness in significantly boosting the performance of existing image classifiers, particularly on hard data.

\end{itemize}


\vspace*{-0.1cm}

%% file: sections/related_works.tex
\section{Related Work}
\vspace*{-0.2cm}
\paragraph{Diffusion models for Synthetic Data}

Recently, a diverse array of generative diffusion models have been proposed, including GLIDE \citep{halgren2004glide}, Imagen \citep{saharia2022photorealistic}, Stable Diffusion \citep{rombach2022high}, Dall-E \citep{ramesh2022hierarchical}, and Muse \citep{chang2023muse}. These models can generate realistic, high-resolution images when conditioned on text prompts, and therefore, are used off-the-shelf to augment the datasets for enhancing data diversity.
For instance, \citet{he2022synthetic} demonstrates that synthetic data created with GLIDE can significantly improve both zero-shot and few-shot performance on image classification.
Recent works like \citet{bansal2023leaving}, \citet{sariyildiz2022fake}, \citet{dunlap2024diversify} and \citet{hemmat2023feedback} have shown that real data combined with synthetic data generated by Stable Diffusion models, boosts the robustness of standard ImageNet classifiers.
Other works such as \citet{diffult2024} train a diffusion model on original data to generate large-scale synthetic samples across the distribution, improving alignment  with real data but limiting diversity. \citet{hemmat2023feedback} further improves diversity by employing an off-the-shelf  diffusion model for single-stage synthetic generation at a large scale.
In contrast, in this work, we focus on learning \textit{hard} data,  and adopt a progressive approach in generating only-useful synthetic data at a significantly smaller scale. We also leverage an off-the-shelf diffusion model, but unlike prior works, we harness different image guidance levels to generate training samples at each stage of training. This method allows for smooth transition across spectrum of interpolations from syn-to-real data, and adapts the model to diverse, in-the-wild scenarios, while maintaining data diversity and alignment.
\vspace*{-0.3cm}
\paragraph{Curriculum Learning (CL)}
Curriculum Learning (CL) was first proposed by \citet{bengio2009curriculum}, introducing a training method analogous to the step-by-step progressive learning of humans. Subsequent works have further explored this idea; for example, \citet{jiang2015self, zhou2020DIH} adjusted the progression pace based on the difficulty of samples, and \citet{jiang2014self, zhou2018scheduled} further take the data diversity into account.  
Previous works \citep{guo2018curriculumnet, zhou2021robust, yuan2022easy} have tried CL on more challenging domains like noisy web images and visual QA; this highlights its potential in tackling challenging scenarios. Few works have explored the combination of data augmentation and curriculum learning~\citep{Hou_2023_ICCV}, but mainly for the text data \citep{pcc_2023, EfficientCL_2021}. Some initial efforts have been made by \citet{ahn2023cuda} to combine CL with engineered image augmentations for tail classes in long-tail learning. In contrast, our work aims to design a generative curriculum on a syn-to-real spectrum of data produced by diffusion models, with broader applications in learning from long-tail or low-quality data.

\vspace*{-0.1cm}

%% file: sections/sec_3_method.tex
\section{Methodology}
\vspace*{-0.2cm}

\input{Figure_tex/teasor_fig}

We propose diffusion curriculum (DisCL) to \textit{``close the distribution gap between original data and the target data distribution''}. DisCL comprises two phases: (Phase 1) Synthetic-to-Real Data Generation that generates a syn-to-real spectrum of interpolated data for hard samples, and (Phase 2) Generative Curriculum learning based on the synthetic data from Phase 1. The two phases are illustrated in Fig.~\ref{fig:teaser}. \looseness-1

\vspace*{-0.1cm}
\subsection{Synthetic-to-Real Data Generation}
\label{sec:s2r}
\vspace*{-0.1cm}
\paragraph{Hard Sample Identification}
We first identify the difficult samples where the model struggles to extract helpful features for target classification. The difficulty estimation can be task-specific. For instance, in long-tail classification with scarce data, the difficulty of each sample depends on whether it belongs to tail classes. For tasks with low-quality data, we can utilize the loss or confidence on the ground-truth class to measure the difficulty.
These samples are marked as ``hard samples'' within the training set (see Fig.~\ref{fig:teaser}), to highlight their role in the model's learning process.

\vspace*{-0.3cm}
\paragraph{Synthetic Data Generation with Image Guidance}
\input{Figure_tex/fig2_sample_img}
Classifier-free guidance in diffusion models was  introduced by \citet{ho2022classifier}, to integrate conditional information into the image denoising process of diffusion without requiring a classifier. It has been adopted by several Text-to-Image generation models such as Stable Diffusion (SD) \citep{rombach2022high}. Given the latent representation of original image as $z_{real}$, the denoising (backward diffusion) process can start from any step $t$ with initial $z_t$ defined as:
\begin{equation}
\vspace*{-0.3cm}
\label{eq:denoisingstep_startingpoint}
    \begin{split}   
        z_t &= \sqrt{\Tilde{\alpha}_t} z_{real} + \sqrt{1 - \Tilde{\alpha}_t}\boldsymbol{\epsilon},~\boldsymbol{\epsilon}\sim\mathcal N(0,\boldsymbol{I}). 
    \end{split}
\end{equation}
The remaining denoising steps iteratively apply the following process of noise estimation  $\hat{\boldsymbol{\epsilon}}_t$ at each step $t$ to get a less noisy generation of $z_{t-1}$, until $t=0$, resulting in a synthetic image $z_0$.  
\vspace*{-0.3cm}
\begin{equation}
\label{eq:denoisingstep}
\centering
    \begin{split}
    \hspace{-.5em}
        \hat{\boldsymbol{\epsilon}}_t = (1 + w) \boldsymbol{\epsilon}_{\theta} (z_t, t|c) - w \boldsymbol{\epsilon}_{\theta}(z_t, t),\\
        z_{t-1} = \frac{1}{\sqrt{\alpha_t}}\left( z_t - \frac{\beta_t}{\sqrt{1 - \Tilde{\alpha}}_t} \hat{\boldsymbol{\epsilon}}_t \right) + \sqrt{\beta_t} \boldsymbol{\epsilon}',~~
        t\leftarrow t-1
    \end{split}
\end{equation}
In Eq.~\ref{eq:denoisingstep_startingpoint}-\ref{eq:denoisingstep}, $\Tilde{\alpha}_t,\alpha_t,$ and $\beta_t$ together define the variance schedule of the diffusion process.
$\boldsymbol{\epsilon}, \boldsymbol{\epsilon}'\sim\mathcal{N}(0, \boldsymbol{I})$ are two independently-sampled Gaussian noises, $\boldsymbol{\epsilon}_{\theta}(\cdot,\cdot)$ refers to the noise estimation model, 
and \(w \in \mathcal{R}\) controls the strength of the textual prompt \(c\) as a condition to $\boldsymbol{\epsilon}_\theta(\cdot,\cdot)$. 

Since $\Tilde{\alpha}_t$ monotonically decreases with $t$, the choice of the initial $t$ in Eq.~\ref{eq:denoisingstep_startingpoint} controls the impact of the original $z_{real}$ in the denoising process, and more visual information of $z_{real}$ tends to be preserved in \(z_0\) if initializing from a small $t$. 
To achieve a full spectrum of interpolations between the real image $z_{real}$ and synthetic images depicted by $c$, following prior work in \citet{meng2021sdedit}, we modify the initial step $t$ in Eq.~\ref{eq:denoisingstep_startingpoint} to $t(\lambda)\triangleq \lfloor (1-\lambda) T \rfloor$ where \(\lambda \in [0, 1)\) defines the image-guidance level, i.e., 
\begin{equation}
\label{eq:imageguidance}
    z_{t(\lambda)} = \sqrt{\tilde{\alpha}_{t(\lambda)}} z_{real} + \sqrt{1 - \tilde{\alpha}_{t(\lambda)}} \boldsymbol{\epsilon},~~t(\lambda)\triangleq \lfloor (1-\lambda) T \rfloor.
\end{equation}
Hence, a larger guidance level \(\lambda\) leads to higher fidelity of generated image $z_0$ to original $z_{real}$, while a smaller \(\lambda\) results in a more prototypical image $z'_0$ depicted by textual prompt $c$.
\(\lambda=0\) results in a generated image based on text only. \footnote{$\lambda = 0$ corresponds to images generated using only text prompt guidance, while $\lambda = 1$ corresponds to replication of the original image without any text guidance.}
\looseness-1
\vspace*{-0.3cm}
\paragraph{Synthetic-to-Real Spectrum of Generated Images}
We use state-of-the-art Stable Diffusion Model \footnote{We use Stable Diffusion XL model for generation} to generate synthetic images for the hard samples identified in Phase 1 of Fig.~\ref{fig:teaser}. 
By adjusting the image guidance scale \(\lambda \in [0, 1)\) in Eq.~\ref{eq:imageguidance}, the denoising process in Eq.~\ref{eq:denoisingstep} can produce a full spectrum of smooth transitions between text-only guided synthetic images and real images. 
We next study the effect of varying the image guidance scales \(\lambda\) on the generated synthetic images. As shown in Fig.~\ref{fig:SyntheticIntepolation1}, changing \(\lambda\) leads to varying difficulty and diversity of synthetic images.
With a smaller \( \lambda \), diffusion model mainly relies on the text information provided in the prompt $c$, generating synthetic images that differ markedly from the original and focus more on the distinct prototypical features of the class in $c$.
As \( \lambda \) increases, the synthetic images increasingly inclines towards the original image, exhibiting less diversity (across random seeds) and more resemblance to the original ones. When the original images are of low-quality, 
a large \( \lambda \) makes it challenging for the classifier to learn discriminating features from synthetic images. 
Therefore, the broad spectrum of synthetic data offers diverse properties, e.g., diversity, hardness, proximity to the real ones, providing a \textit{novel} design space for curriculum learning.\looseness-1

\vspace*{-0.3cm}
\paragraph{Filter out Synthetic Data with Low-Fidelity}
As shown in Fig.~\ref{fig:SyntheticIntepolation1}, some synthetic images may suffer from poor quality and low fidelity to the text prompt $c$, \textit{e.g.} the class object is missing or obscured, which can hinder the downstream tasks. To mitigate this, we filter-out such images by applying CLIP-based filtering used in \citep{clipscore2022, schuhmann2021laion, dunlap2024diversify}, which measures CLIP cosine similarity between synthetic images and text prompt. 
We discard images that fall below a predefined CLIPScore threshold before the training begins. \looseness-1

\vspace*{-0.1cm}
\subsection{Generative Curriculum with Synthetic Data}
\label{sec:introduce_curr_method}
\vspace*{-0.17cm}
With the full spectrum of syn-to-real generated data, we achieve a smooth transition from images with prototypical features and high diversity to task-specific features that closely resemble real images. This enables us to design a curriculum that selects data based on  their 
diversity and feature types for different training stages. 
With a curriculum of rich synthetic data, we can improve the model's performance in challenging and diverse cases that would be difficult to address using only real data. Additionally, this approach allows us to control the distribution gap to the original data.
We apply our generative curriculum to two challenging applications: long tail learning and learning from low-quality.
\vspace*{-0.8cm}
\paragraph{For \textbf{long tail classification},}
\label{para:intro_LT}
the scarcity of data in minority classes makes it difficult for models to extract useful features for these classes, leading to poor generalization on balanced test set. For tail classes, we first generate a full spectrum of synthetic data using techniques in \S\ref{sec:s2r}, following the standard split of tail classes in the studied dataset. A diverse set of textual prompts is used to achieve this goal\footnote{Text prompts are provided in Appendix~\ref{sec:app_inltprompt}}. The generated spectrum offers varying degrees of data diversity, which if used at once, can introduce a syn-to-real distribution gap. To mitigate this gap, we first expose the model to diverse synthetic images of tail classes, and then progressively shift to a task-specific distribution that resembles the original images. This yields a \textit{non-adaptive} \textbf{``Diverse-to-Specific'' curriculum} that starts with synthetic data with a lower guidance scale ($\lambda \rightarrow$  0) and gradually moves toward data of a higher guidance scale ($\lambda \rightarrow$  1). The algorithm for our \textit{non-adaptive} curriculum strategy is provided here in Algorithm~\ref{alg:DisCLnonadaptive}.
\vspace*{-0.3cm}

\paragraph{\textbf{Learning from low-quality}} images
\label{para:intro_wild}
can be challenging due to inevitable quality issues, such as obscurity in images from traffic, motion, or wildlife observation cameras. We investigate wildlife observation as an example application of DisCL to enable effective learning under such scenarios. For low-quality camera trap images, we aim to generate simpler, more prototypical images of animals to warm up training and generalize to harder cases. We identify hard samples using a pretrained classifier—lower ground-truth class probabilities indicate higher difficulty. Varying the image guidance scale, we synthesize a full spectrum of data for these samples using class information in the text prompts\footnote{Text prompts are provided in Appendix~\ref{sec:app_wildprompt}.}, which steers the diffusion model toward relevant animal and habitat features. Unlike long-tail settings, hard samples in low-quality domains often lack both prototypical and generalizable features, as shown in prior work on camera trap images \citep{ml-for-camera-trap-images}. Consequently, synthetic data generated with low guidance often appears prototypical but out-of-distribution. A non-adaptive curriculum that introduces such synthetic data early risks distribution shift or overemphasis on outlier features. DoCL \citep{zhou2021curriculum} addressed this by selecting real data adaptively to optimize learning progress. Inspired by this, we propose an \textit{adaptive curriculum} detailed in Algorithm ~\ref{alg:DisCLadaptive}, which selects the image guidance level $\lambda$ at each epoch based on progress -- defined by improvement in ground-truth class confidence on validation subsets corresponding to each $\lambda$. The guidance level with the highest progress is chosen for the next epoch. This ensures the model learns from the most informative data at each stage, enabling a smooth transition from simple to complex patterns and maximizing improvement on the \textit{real data} distribution. Details are in Algorithm~\ref{alg:DisCLadaptive}.
\vspace*{-0.1cm}

%% file: Figure_tex/teasor_fig.tex
\begin{figure*}[ht]
  \centering
\includegraphics[width=0.9\textwidth]{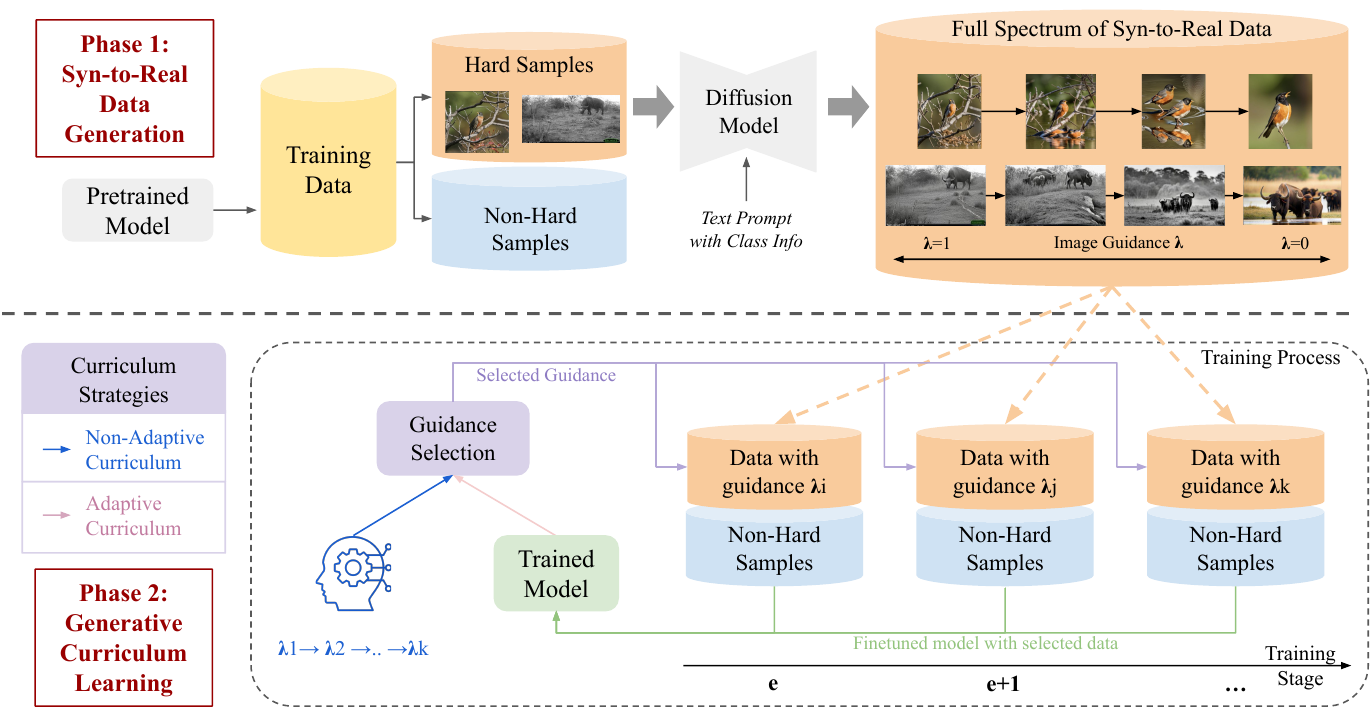}
\caption{\textbf{Overview of Diffusion Curriculum (DisCL)}. DisCL consists of two phases: (Phase 1) Syn-to-Real Data Generation and (Phase 2) Generative Curriculum learning. 
In Phase 1, we use a pretrained model to identify the ``hard'' samples in the original images and use them as guidance to generate a full spectrum of synthetic to real images by varying image guidance strength \(\lambda\). 
In Phase 2, a curriculum strategy (Non-Adaptive or Adaptive) selects training data from the full spectrum, by determining image guidance level $\lambda_i$ for each training stage $e$. The Adaptive strategy chooses $\lambda_i$ to maximize expected progress, while the Non-Adaptive strategy follows a predefined schedule. Synthetic data of the selected guidance level is then combined with non-hard real samples to train the task model.}
\label{fig:teaser}
\vspace*{-1\baselineskip}
\end{figure*}

%% file: Figure_tex/fig2_sample_img.tex
\begin{figure*}[ht]
\vspace*{-0.09cm}
\centering
\includegraphics[width=0.7\textwidth]{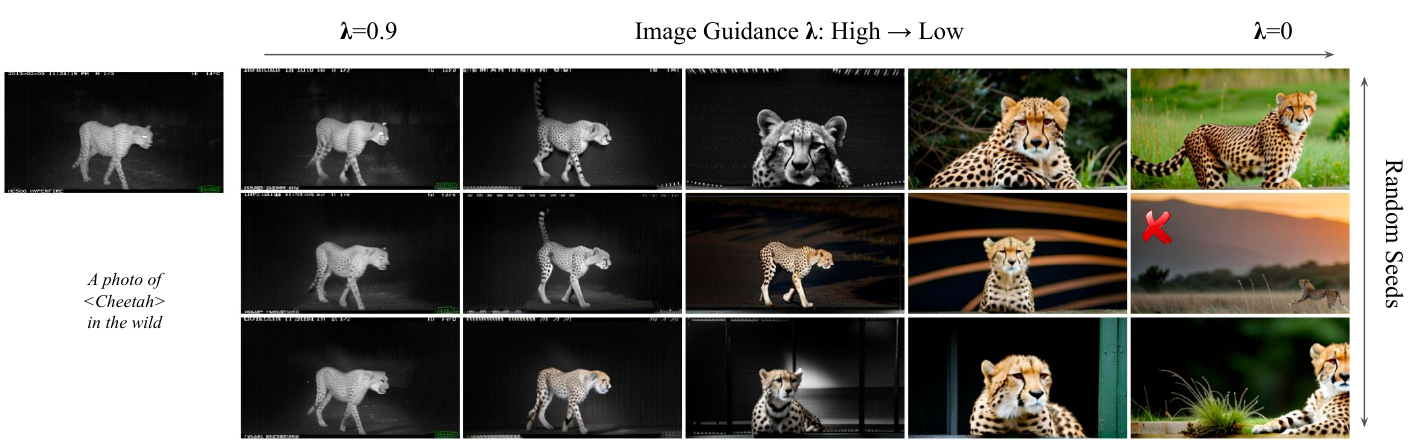}
\includegraphics[width=0.7\textwidth]{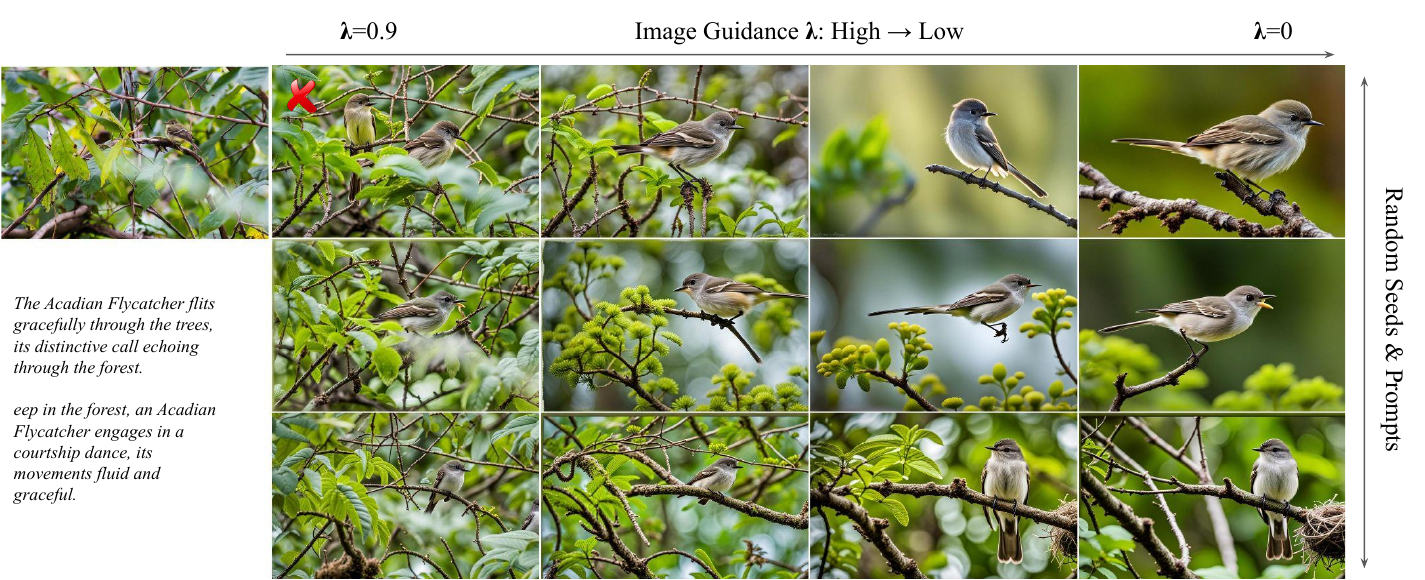}
  \caption{Synthetic images generated with different image guidance levels $\lambda$ and random seeds. \textcolor{red}{\textbf{$\boldsymbol{\times}$}} marks images with low-fidelity to the text prompt, which are filtered out by CLIPScore (ref. the end of \S\ref{sec:s2r}). Zoom-in for better view. \looseness-1}
    \label{fig:SyntheticIntepolation1}
\vspace*{-1\baselineskip}
\end{figure*}

%% file: sections/experiments.tex
\section{Experiments}
\label{sec:experiments}
\vspace*{-0.15cm}
\subsection{Long-Tail Classification}
\vspace*{-0.1cm}
\paragraph{Setup} \label{sec:inltsetup}
To validate the efficacy of DisCL method on long-tail classification, we conduct main experiments with ImageNet-LT (IN-LT) dataset \citep{liu2019large}. This dataset includes 1000 classes, with class cardinality ranging from 5 to 1,280. To assess the robustness of DisCL more comprehensively, we conduct experiments on two additional datasets: a synthetically imbalanced dataset, CIFAR100-LT \citep{cao2019learning}, and a real-world benchmark, iNaturalist2018 \citep{van2018inaturalist}. CIFAR100-LT is provided with imbalanced classes by synthetically sampling the training data with multiple imbalance ratios $\{100, 50\}$. iNaturalist2018 dataset represents a naturally occurring long-tailed distribution with class cardinality ranging from 2 to 1000. 
We evaluate overall accuracy and the accuracy across three class categories: many (most frequent), medium, and few (least frequent, tail) classes on the standard balanced test sets of three datasets.
For synthetic data generation, we use DDIM \citep{song2020denoising} as our noise scheduler.
For training, we follow \citet{ahn2023cuda} and \citet{han2024latent}, using ResNet-10 as visual backbone. We average results over 3 runs and report training details and hyper-parameters in Appendix~\ref{sec:app_inltimplementation} and ~\ref{sec:app_hyper}. 
\vspace*{-0.3cm}
\paragraph{Baselines}
We compare the effect of DisCL with comparable baseline of CUDA \citep{ahn2023cuda} and LDMLR \citep{han2024latent}, mainly using Cross-Entropy (CE) loss function. To further illustrate the robustness of DisCL, we try Balanced Softmax (BS) loss \citep{bs2020}, known for its competitive performance on long-tail learning. 
\begin{itemize}[leftmargin=*]
    \label{itemize:inltbaseline}
    \item \textbf{CUDA}: Engineered data augmentation + curriculum learning on IN-LT. 
    \item \textbf{LDMLR}: A three-stage training using diffusion model to improve LT. 
    \item \textbf{BS loss}: Balanced \texttt{softmax} to address class-distribution shift between training and test sets. \looseness-1
\end{itemize}

We also conduct ablation study to analyze the effect of DisCL under different hyperparameter settings. 
We note that, real data for hard samples ($\lambda\sim$1) is included by default; however, this doesn't apply to the Fixed Guidance and Text-only Guidance ablation:
\begin{itemize}[leftmargin=*]
    \item \textbf{Text-only Guidance}: Using data at image guidance scale \(\lambda = 0\) without curriculum strategy. 
    \item \textbf{Fixed Guidance}
    \footnote{\textit{Text-only Guidance} (\(\lambda\)=\(0\)) reaches the best performance amongst all guidance scales. Hence, the result of  \textit{Fixed Guidance} here are same as  \textit{Text-only Guidance}, reported in Table~\ref{tab:inltmainresults}. We also report the performance of all other scales in \textit{Fixed Guidance} experiment in the Fig.~\ref{fig:GuidanceDifference}.}: uses data generated from a single guidance scale \(\lambda_i \in [0, 1)\). We report results for the guidance with the highest few-class accuracy. 
    \item \textbf{DisCL}: employs multiple levels of guidance scales alongside a range of curriculum strategies. These strategies and the guidance intervals used for training, are defined below:
    \begin{itemize}[leftmargin=*]
        \item \textbf{Diverse to Specific}: Non-adaptive strategy with guidance changing from smallest (diverse augmentation) to largest (task-specific augmentation). 
        \item \textbf{Specific to Diverse}: Non-adaptive strategy with guidance changing from largest to smallest.
        \item \textbf{Adaptive}: Curriculum strategy\footnote{Curriculum strategy proposed in \S\ref{para:intro_wild}} to adaptively select guidance during training. 
    \end{itemize}
\end{itemize}
\vspace*{-0.35cm}
\paragraph{Results}
\input{Table_tex/inlt_mainresults}
We present the results of our method alongside the baselines for the ImageNet-LT dataset in Table~\ref{tab:inltmainresults}. With CE loss, DisCL significantly improves accuracy in all 4 class-categories. Notably, ``Few'' class accuracy increases by 17.06\%, from 6.58\% to 23.64\%, demonstrating DisCL's effectiveness in addressing the data scarcity challenge, especially for tail classes. DisCL also works effectively with BS loss, resulting in additional gains of 1.52\% in Many, 2.08\% in Few, and 1.3\% overall accuracy, further underscoring our method's impact even with a class-balancing loss function.
Results on CIFAR100-LT (Table~\ref{tab:cifar100}) and iNaturalist2018 (Table~\ref{tab:inat2018}) confirm the robustness of DisCL across several diverse datasets, achieving better accuracy in tail classes along with improved overall generalization.
\vspace*{-0.01cm}

\input{Table_tex/cifar_and_inat}

\subsection{Learning from Low-quality Data}
\vspace*{-0.1cm}
\paragraph{Setup}\label{iwild_setup}
We also conduct DisCL experiments with iWildCam dataset \citep{beery2021iwildcam} to evaluate its efficacy in classifying low-quality data. The task is to classify 182 different animal species from images captured by camera traps. 
We evaluate model performance on standard out-of-domain (OOD) and in-domain (ID) test sets in terms of macro F1 score. 
We choose the CLIP ViT model as our base model and finetune CLIP ViT-B/16 and CLIP ViT-L/14 \footnote{We use hyperparameters provided in \citet{goyal2023finetune} with a batchsize of 128 to train the model.} models with DisCL. 
The reported accuracy is averaged over 3 random seeds. More training details and hyperparameters are provided in Appendix~\ref{sec:app_wildimplementation} and Appendix~\ref{sec:app_hyper}. \looseness-1
\vspace*{-0.3cm}
\paragraph{Baselines}
We compare the effect of our method with three benchmark algorithms, LP-FT \citep{kumar2022fine}, FLYP \citep{goyal2023finetune}, and ALIA \citep{dunlap2024diversify}.
To further analyze the gain of our model, we try Weighted Ensembling (WE) method \citep{wortsman2022robust}, which can further improve model performance by integrating prior knowledge from pretrained model: 
\begin{itemize}[leftmargin=*]
    \label{itemize:iwildcambaseline}
    \item \textbf{LP-FT}: A two-step process involving linear probing and full fine-tuning of model to avoid distortion of pretrained features, to improve OOD generalization.
    \item \textbf{FLYP}: Finetuning with the pretraining (contrastive) loss. \looseness-1
    \item \textbf{ALIA}: Diffusion-based data-augmentation on fine-grained classification tasks.\looseness-1
    \item \textbf{WE}: Linearly merging the weights of pretrained and finetuned model. 
\end{itemize}

We conduct ablation study to analyze the effect of DisCL with different hyper-parameters introduced in \S\ref{sec:inltsetup}, and the newly introduced ablation hyper-parameters:
\begin{itemize}[leftmargin=*]
    \item \textbf{DisCL}: employs multiple levels of guidance scale and a range of curriculum strategies:
    \begin{itemize}[leftmargin=*]
        \setlength{\itemsep}{0pt}  
        \setlength{\parindent}{-1pt} 
        \item \hspace*{-0cm}\textbf{Easy to Hard}: Non-adaptive strategy with guidance changing from smallest (easiest \& most prototypical features) to largest (hardest \& task-specific features).
        \item \hspace*{-0cm}\textbf{Random}: Randomly select  guidance at each train stage.
    \end{itemize}
\vspace{-0.2cm}    
\end{itemize}
\vspace{-0.2cm}
\paragraph{Results}
\input{Table_tex/iwildcam_mainresults}
We present the results of our method and comparable baselines for the iWildCam dataset in Table~\ref{tab:iwildcammainresult}. Compared to the nearest competing baseline, DisCL significantly enhances the OOD F1 performance by 2.6\%. Additionally, DisCL boosts the ID F1 performance by 2.1\%. Among all evaluated methods, DisCL achieves the highest scores in both OOD and ID generalization, underscoring its effectiveness for low-quality classification. Moreover, our model could still deliver performance improvements on larger model when using ViT-L/14, as shown in Fig~\ref{fig:iwildcam_clipvit16}; DisCL  achieves gains of 2.8\% in OOD F1 and 3.7\% in ID F1. These findings reinforce the versatility and robustness of the DisCL framework across different model scales and complexities.
We further study the performance of model after employing WE method. DisCL still benefits from this method and maintains superior performance compared to other methodologies, despite integrating prototypical features from synthetic data that might overlap with the pretrained model's knowledge.
\vspace*{-0.1cm}

%% file: Table_tex/inlt_mainresults.tex
\begin{table}[t]
\centering
\resizebox{1.01\columnwidth}{!}{%
\begin{tabular}{@{}c|lc|cccc@{}}
\toprule
&              &      & \multicolumn{4}{c}{\textbf{ImageNet-LT}} \\
& \textbf{Method} & \textbf{Curriculum}     &  \textbf{Many}  & \textbf{Medium} & \textbf{Few} & \textbf{Overall}  \\
\midrule
\parbox[t]{2mm}{\multirow{5}{*}{\rotatebox[origin=c]{90}{Baselines}}} & \small{CE}    &N/A                   & 57.70   & 26.60   & 4.40  & 35.80 \\
 & \small{CE + LDMLR}      &N/A       &  57.20  & 29.20   & 7.30  & 37.20  \\
 & \small{CE + CUDA} &N/A   &  \textbf{57.49}  & 28.16   & 6.58  & 36.30  \\
 & \small{\textcolor{orange}{BS}\(\dagger\)} &N/A    & 51.14 & 37.02   & 19.29  & 39.80 \\
 & \small{\textcolor{orange}{BS} + CUDA\(\dagger\)} &N/A  &  51.16  &  37.35  &  19.28 & 40.03  \\
\midrule
\parbox[t]{2mm}{\multirow{7}{*}{\rotatebox[origin=c]{90}{Ablations}}}  & \small{CE + Text-only Guidance} &N/A  &  56.63  & 30.69 & 17.90  & 39.10 \\
 & \small{CE + All-Level Guidance}         &N/A          & 56.77    & 30.88 & 19.17  & 39.40 \\
 & \small{CE + DisCL}    &  Adaptive    &  56.21  &  30.43  & 16.78  & 38.65  \\
 & \small{CE + DisCL}  &  Specific to Diverse      &  56.71  &  30.67  & 18.36  & 39.18  \\
 & \small{CE + DisCL [Lower CLIPScore Threshold]}  &  Diverse to Specific       &  57.66  &  30.61  & 23.69  & 39.67  \\
 & \small{CE + DisCL [Higher CLIPScore Threshold]}   &  Diverse to Specific      &  56.92  & 30.64  & 22.88  &  39.68 \\
\midrule
\parbox[t]{2mm}{\multirow{2}{*}{\rotatebox[origin=c]{90}{Ours}}} & \small{CE + DisCL}    &  Diverse to Specific   &  56.78   &  \textbf{30.73}  & \textbf{23.64}  & \textbf{39.82}   \\
 & \small{\textcolor{orange}{BS} + DisCL} &  Diverse to Specific &  \textbf{52.68}  & \textbf{37.68}   & \textbf{21.36}  &  \textbf{41.33}\\
\bottomrule
\end{tabular}%
}
\vspace*{-0.2cm}
\caption{
\small{Accuracy (\%) of long-tail classification on ImageNet-LT with base model ResNet-10. The best accuracies among baseline and DisCL are highlighted in \textbf{bold}. \(\dagger\) marks our reproduced results using the original paper-provided code. \textcolor{orange}{BS} refers to Balanced Softmax Loss\citep{bs2020}. Baselines (LDMLR, CUDA) are defined in \S\ref{itemize:inltbaseline}.}\looseness-1}
\vspace*{0.05cm}
\label{tab:inltmainresults}
\vspace*{-1\baselineskip}
\end{table}

%% file: Table_tex/cifar_and_inat.tex
\begin{table}[ht]
\centering
\resizebox{\linewidth}{!}{%
\begin{tabular}{lc|cccc|cccc}
\toprule
& & \multicolumn{8}{c}{\textbf{CIFAR-100-LT}} \\
& & \multicolumn{4}{c|}{\textbf{Imbalance Ratio=100}} & \multicolumn{4}{c}{\textbf{Imbalance Ratio=50}} \\
\textbf{Method}  & \textbf{Curriculum}   &  \textbf{Many}  & \textbf{Medium} & \textbf{Few} & \textbf{Overall} &  \textbf{Many}  & \textbf{Medium} & \textbf{Few} & \textbf{Overall} \\
\midrule
\small{CE}        &N/A             & 52.86 & 25.34 & 5.49 & 29.02 & 49.60 & 25.41 & 5.33 & 31.72 \\
\small{CE + CUDA} &N/A & \textbf{54.55} & \textbf{26.07} & 5.43 & 29.85 & 52.29 & 26.17 & 5.53 & 33.13  \\
\textbf{\small{CE + DisCL}}   &  Diverse to Specific   & 53.14 & 25.52 & \textbf{10.65} & \textbf{30.91} & \textbf{53.4} & \textbf{31.69} & \textbf{21.47} & \textbf{36.22} \\
\midrule
\small{\textcolor{orange}{BS}}          &N/A             & 47.87 & 30.07 & 14.41 & 31.61 & 46.01 & 30.76 & 18.55 & 34.82 \\
\small{\textcolor{orange}{BS} + CUDA} &N/A & 48.01 & \textbf{32.79} & 15.55 & 33.02 & 46.08 & 32.51 & 22.11 & 36.21  \\
\textbf{\small{\textcolor{orange}{BS} + DisCL}}  &   Diverse to Specific   & \textbf{49.02} & 29.02 & \textbf{19.07} & \textbf{33.08} & \textbf{49.51} & \textbf{32.60} & \textbf{25.58} & \textbf{36.77} \\
\bottomrule
\end{tabular}%
}
\vspace*{-0.2cm}
\caption{\small{Accuracy (\%) of long-tail classification on CIFAR-100-LT with base model ResNet-10. The best accuracy for overall and classes of \{many, medium, few\} samples is highlighted in \textbf{bold}.}}
\vspace*{0.05cm}
\label{tab:cifar100}
\vspace*{-0.1\baselineskip}
\end{table}

\begin{table}[ht]
\centering
\resizebox{0.8\linewidth}{!}{%
\begin{tabular}{lc|cccc}
\toprule
& & \multicolumn{4}{c}{\textbf{iNaturalist2018}} \\
\textbf{Method}  & \textbf{Curriculum}   &  \textbf{Many}  & \textbf{Medium} & \textbf{Few} & \textbf{Overall} \\
\midrule
\small{CE}        &N/A             & 55.02 & 43.40 & 37.33 & 42.20 \\
\small{CE + CUDA} &N/A & \textbf{55.94} & 44.21 & 39.13 & 43.18 \\
\textbf{\small{CE + DisCL}}   &  Diverse to Specific   & 54.71 & \textbf{44.37} & \textbf{48.92} & \textbf{47.25} \\
\midrule
\small{\textcolor{orange}{BS}}          &N/A             & 46.12 & 49.31 & 50.27 & 49.46 \\
\small{\textcolor{orange}{BS} + CUDA} &N/A & \textbf{48.77} & \textbf{49.94} & 50.87 & 50.23 \\
\textbf{\small{\textcolor{orange}{BS} + DisCL}}  &   Diverse to Specific   & 45.44 & 48.18 & \textbf{53.63} & \textbf{50.30} \\
\bottomrule
\end{tabular}%
}
\vspace*{-0.2cm}
\caption{\small{Accuracy (\%) of long-tail classification on iNaturalist2018 with base model ResNet-10. The best accuracy for overall and classes of \{many, medium, few\} samples is highlighted in \textbf{bold}.}}
\vspace*{0.05cm}
\label{tab:inat2018}
\vspace*{-1\baselineskip}
\end{table}

%% file: Table_tex/iwildcam_mainresults.tex
\begin{table}[ht]
\centering
\resizebox{\columnwidth}{!}{%
\begin{tabular}{@{}c|lc|cc@{}}
\toprule
 &          &        & \multicolumn{2}{c}{\textbf{iWildCam}} \\
 & \textbf{Method}  & \textbf{Curriculum}     & \textbf{OOD}    & \textbf{ID}  \\
\midrule
\parbox[t]{2mm}{\multirow{5}{*}{\rotatebox[origin=c]{90}{Baselines}}} & \small{CLIP (zero-shot)}                &        & 11.0 (-)   & 8.7 (-)    \\
 & \small{LP-FT}         &N/A                           & 34.7 (0.4)    & 49.7 (0.5)    \\
 & \small{LP-FT + WE}  &N/A     &  35.7 (0.4)  &   50.2 (0.5)   \\
 & \small{FLYP\(\dagger\)}       &N/A                    & 35.5 (1.1)    & 52.2 (0.6)    \\
 & \small{FLYP + WE\(\dagger\)}   &N/A    &  36.4 (1.2)  &   52.0 (1.0)   \\
 & \small{FLYP + ALIA}       &N/A                    & 36.9 (0.3)    & 52.6 (0.4)    \\
\midrule
\parbox[t]{2mm}{\multirow{8}{*}{\rotatebox[origin=c]{90}{Ablations}}}  & \small{FLYP + Text-only Guidance} &N/A & 34.2 (0.4)      & 51.4 (0.3)      \\ 
 & \small{FLYP + Fixed Guidance}  &N/A &   36.0 (0.3)   & 50.8 (0.6)       \\ 
 & \small{FLYP + All-Level Guidance}   &N/A   & 36.5 (0.6)    & 53.4 (0.5)    \\
 & \small{FLYP + DisCL}      & Easy-to-Hard    & 35.2 (0.9)      & 51.4 (0.5)      \\ 
 & \small{FLYP + DisCL}   & Random     & 35.9 (0.1)    & 52.1 (0.2)    \\
 & \small{{FLYP + DisCL [Lower CLIPScore Threshold]}}  & Adaptive &  37.1 (0.8) & 50.9 (0.9) \\
 & \small{{FLYP + DisCL [Higher CLIPScore Threshold]}}  & Adaptive &  38.1 (1.3) & 52.8 (0.8) \\
\midrule
\parbox[t]{2mm}{\multirow{2}{*}{\rotatebox[origin=c]{90}{Ours}}} & \small{{FLYP + DisCL}}  & Adaptive  & \textbf{38.2 (0.5)} & \textbf{54.3 (1.4)}    \\
 & \small{FLYP + DisCL + WE}   & Adaptive    &  \textbf{38.7 (0.4)}  &   \textbf{54.6 (0.7)}   \\
\bottomrule
\end{tabular}
}%
\vspace*{-0.2cm}
\caption{\small{In-distribution (ID) and out-of-distribution (OOD) macro F1 score of low-quality image learning on iWildCam with CLIP ViT-B/16 model. The best performance is highlighted in \textbf{bold}. \(\dagger\) marks our reproduced results using the original paper provided code. Baselines are defined in \S\ref{itemize:iwildcambaseline}.}}
\label{tab:iwildcammainresult}
\vspace*{-1\baselineskip}
\end{table}

%% file: sections/ablation_analysis.tex
\section{Ablation Study and Analysis }
\input{Figure_tex/clip_vit_iwildcam}
\vspace*{-0.15cm}
\subsection{Effect of Syn-to-Real Interpolation Data}
\vspace*{-0.1cm}
We examine the effectiveness of using a spectrum of data generated with our DisCL method, by comparing \textit{All-Level Guidance} and \textit{Text-only Guidance} rows in both the task tables (IN-LT and iWildCam). For IN-LT results in Table~\ref{tab:inltmainresults}, \textit{All-Level Guidance} brings $\sim$1.27$\%$ gain in few-class accuracy, alongwith significant gains across other class-categories. Likewise, \textit{All-Level Guidance} shows a superior ID and OOD performance as compared to \textit{Text-only Guidance} for the iWildCam as well, see Table~\ref{tab:iwildcammainresult}. These findings corroborate that utilizing a spectrum of data with multiple guidance levels helps mitigate the negative effects of the distribution gap.
\vspace*{-0.1cm}
\subsection{Effect of Curriculum Learning Strategy}
\vspace*{-0.1cm}
\paragraph{Long Tail Classification}
We compare the impact of our \textit{Diverse to Specific} curriculum strategy tailored for IN-LT task against other strategies, notably \textit{All-Level Guidance} which employ no curriculum and uses all synthetic data. The \textit{Diverse to Specific} demonstrate a higher few-class accuracy with a margin of 4.47$\%$, see Fig.~\ref{fig:inltcurriculum}. We then compare it with a reverse strategy \textit{Specific-to-Diverse}, and found the latter one to be worse. The reverse strategy can overfit model to real distribution early on, increasing the gap between real and synthetic data; hence, later-stage training on the data with larger distribution gap can decrease models' few-class accuracy. For IN-LT, we also try \textit{Adaptive} strategy (mainly developed for learning from low-quality data), in which strategy's progression is based on a validation set, comprising few tail images sampled from each guidance scale and few original images. 
But, validation set is scarce interms of tail samples, which renders it ineffective for identification of truly useful guidance. Hence, this strategy ranks as the least effective for LT task.

\vspace*{-0.3cm}
\paragraph{Learning from Low Quality Data}
For iWildCam task, we study the effect of our designed \textit{Adaptive} strategy, catering to the challenge of learning from low quality data. As shown in Fig.~\ref{fig:iwild_curr}, for this task, \textit{Adaptive} surpasses the \textit{All-Level Guidance} with a clear margin, underscoring the benefit of using progressive curriculum over using all synthetic data. Further comparisons with the Non-Adaptive curricula including \textit{Easy-to-Hard} and \textit{Random}, show an impactful increase in OOD F1, while using our \textit{Adaptive}.

These findings highlight how the structured data selection used in \textit{Diverse-to-Specific}, is more effective in directing model's focus on scarce data (classes), however, when dealing with real-world low-quality data, an \textit{Adaptive} strategy is more successful in adjusting to models' needs by adaptively selecting the suited data.

\vspace*{-0.1cm}
\subsection{Effect of CLIPScore Threshold }
\vspace*{-0.1cm}
\paragraph{Long Tail Classification}
Our analysis of CLIPScore distribution on IN-LT generated data leads us to infer that the best CLIPScore threshold for filtering is 0.3 (detailed explained in the Appendix~\ref{sec:app_inltfilter}). We then assess different CLIPScore thresholds with the \textit{Diverse to Specific} curriculum strategy, by experimenting with different values: lower (0.28), and higher (0.32), shown in Fig.~\ref{fig:inltinterval}. 
However, we find that changing the CLIPScore threshold does not significantly affect the performance. As shown in Fig.~\ref{fig:textsimilarity_inlt}, CLIPScore of synthetic data remains concentrated, as Stable Diffusion model performs well on generating high-quality images for ImageNet classes. Changes in the CLIPScore threshold will not significantly affect the quality of synthetic images and corresponding effects in downstream classification tasks. 

\vspace*{-0.3cm}
\paragraph{Learning from Low Quality Data}

In the iWildCam task, we identify the optimal threshold as 0.25. To further validate this choice, we experiment with nearby thresholds (0.23 and 0.27) with the chosen \textit{Adaptive Curriculum} strategy suited for low-quality image classification. As depicted in Fig.~\ref{fig:wildinterval}, the 0.25 threshold markedly improves OOD performance compared to other CLIPScore thresholds. Unlike the ImageNet dataset, the iWildCam images are characterized by significant difficulty and poor quality, leading to high variance in CLIPScores of synthetic data (as shown in Fig.~\ref{fig:textsimilarity_wilds}). In this scenario, adjusting the CLIPScore threshold can impact model performance. When a higher threshold is used, the selected synthetic images include more prototypical visual features but they are less similar to the original images. Hence, they improve OOD performance but lead to a drop of ID F1 score. \looseness-1 

The ablation study results on two classification tasks demonstrate that the selection of the CLIPScore threshold should be carefully aligned with the generation quality inherent to the task-at-hand.

\input{Figure_tex/ablation_hyper}

\subsection{Scaling Synthetic Data for Long Tail Learning}
\vspace*{-0.1cm}
We empirically analyze the effect of scaling synthetic tail data on IN-LT performance using ResNet-10 model in Fig.\ref{fig:scale_r10}. 
DisCL  consistently improves the  few class accuracy upto 3-4X scale
; however, beyond this point, the gains diminish with a slight degradation in both many and medium classes. As a result, we chose to use ~3X scale of synthetic tail data for all DisCL training experiments.
Notably, many-class accuracy shows the lowest degradation across scaling, confirming the findings of long-tail learning that hard-sample synthetic data can improve tail class generalization without disrupting many-class representations.
\input{Figure_tex/ablation_scale_synR10}
\vspace*{-0.1cm}

%% file: Figure_tex/clip_vit_iwildcam.tex
\begin{figure}[t]
    \centering
        \centering
        \includegraphics[width=\linewidth]{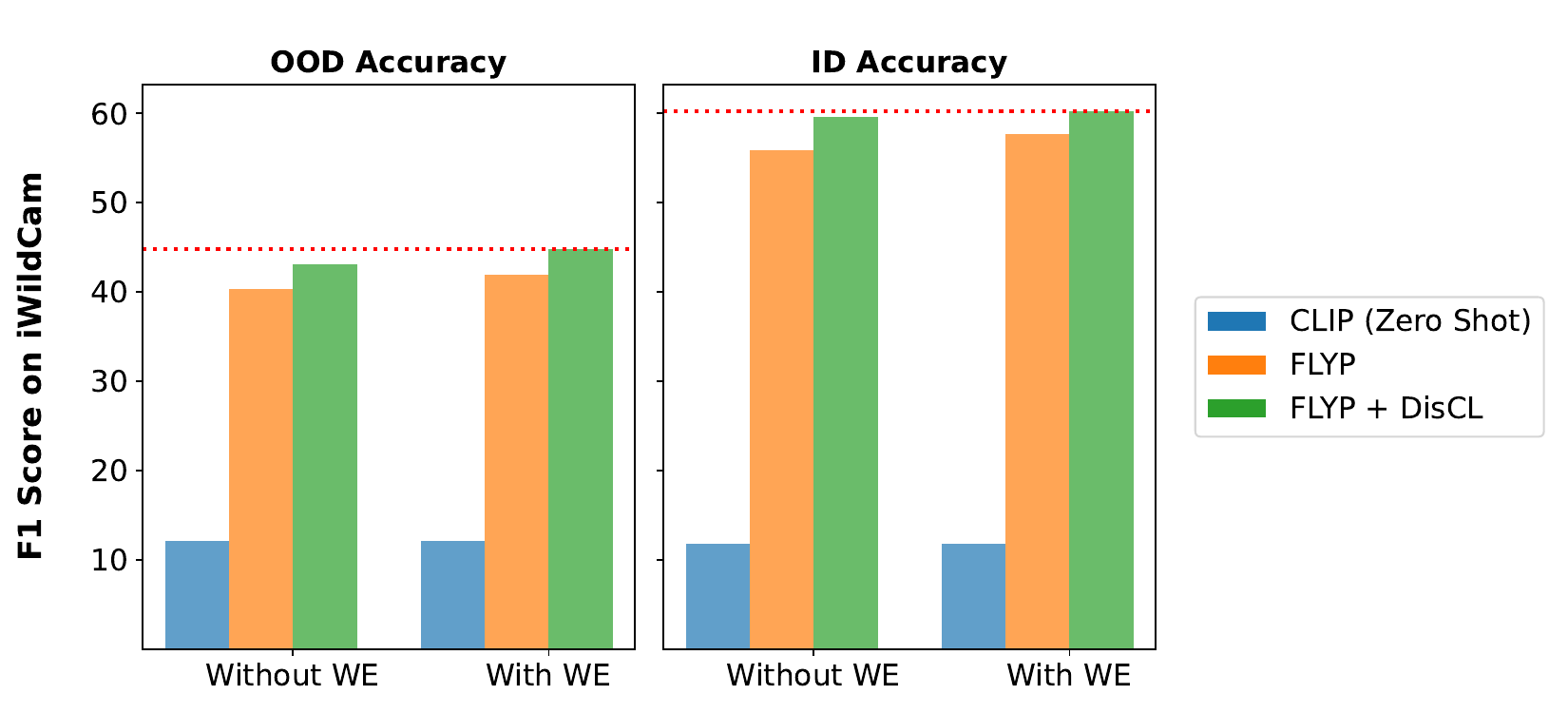}
        \caption{iWildCam accuracy of CLIP ViT-L/16 model trained with and without weight ensembling (WE). The best model performance is highlighted with a \textcolor{red}{red} horizontal line.}
    \label{fig:iwildcam_clipvit16}
    \vspace*{-1.15\baselineskip}
\end{figure}

%% file: Figure_tex/ablation_hyper.tex
\begin{figure}[t]
    \centering
    \begin{subfigure}{0.48\linewidth}
        \centering
        \includegraphics[width=\linewidth]{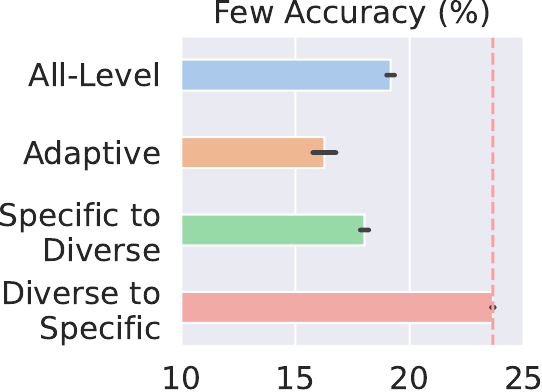}
        \caption{Strategies: IN-LT}
        \label{fig:inltcurriculum}
    \end{subfigure}
    \hfill 
    \begin{subfigure}{0.48\linewidth}
        \centering
        \includegraphics[width=\linewidth]{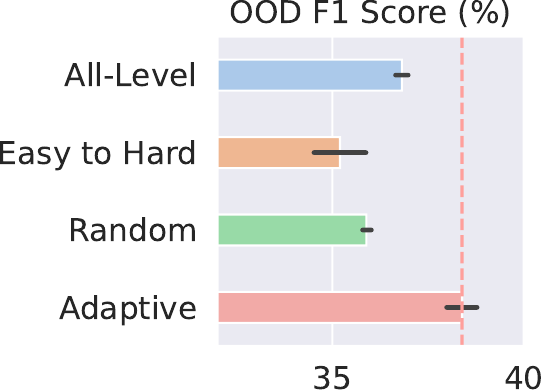}
        \caption{Strategies: iWildCam}
        \label{fig:iwild_curr}
    \end{subfigure}%
    \vspace{0.6em} 
    \begin{subfigure}{0.47\linewidth}
        \centering
        \includegraphics[width=\linewidth]{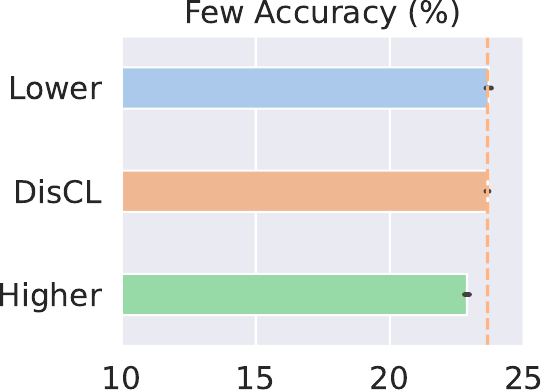}
        \caption{Thresholds: IN-LT}
        \label{fig:inltinterval}
    \end{subfigure}
    \hfill
    \begin{subfigure}{0.47\linewidth}
        \centering
        \includegraphics[width=\linewidth]{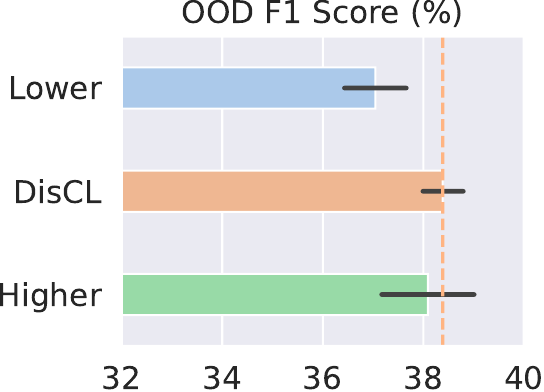}
        \caption{Thresholds: iWildCam}
        \label{fig:wildinterval}
    \end{subfigure}    
    \caption{\small{Ablation study of Curriculum Strategies (a,b) and CLIPScore Thresholds (c,d) $\&$  on ImageNet-LT and iWildCam respectively. Error bar reports the standard deviation of each experiment.}}
    \vspace*{-1\baselineskip}
\end{figure}

%% file: Figure_tex/ablation_scale_synR10.tex
\begin{figure}[t]
    \centering
    \begin{subfigure}{1\linewidth}
        \centering
        \includegraphics[width=\textwidth, trim=10 9 10 9 clip]{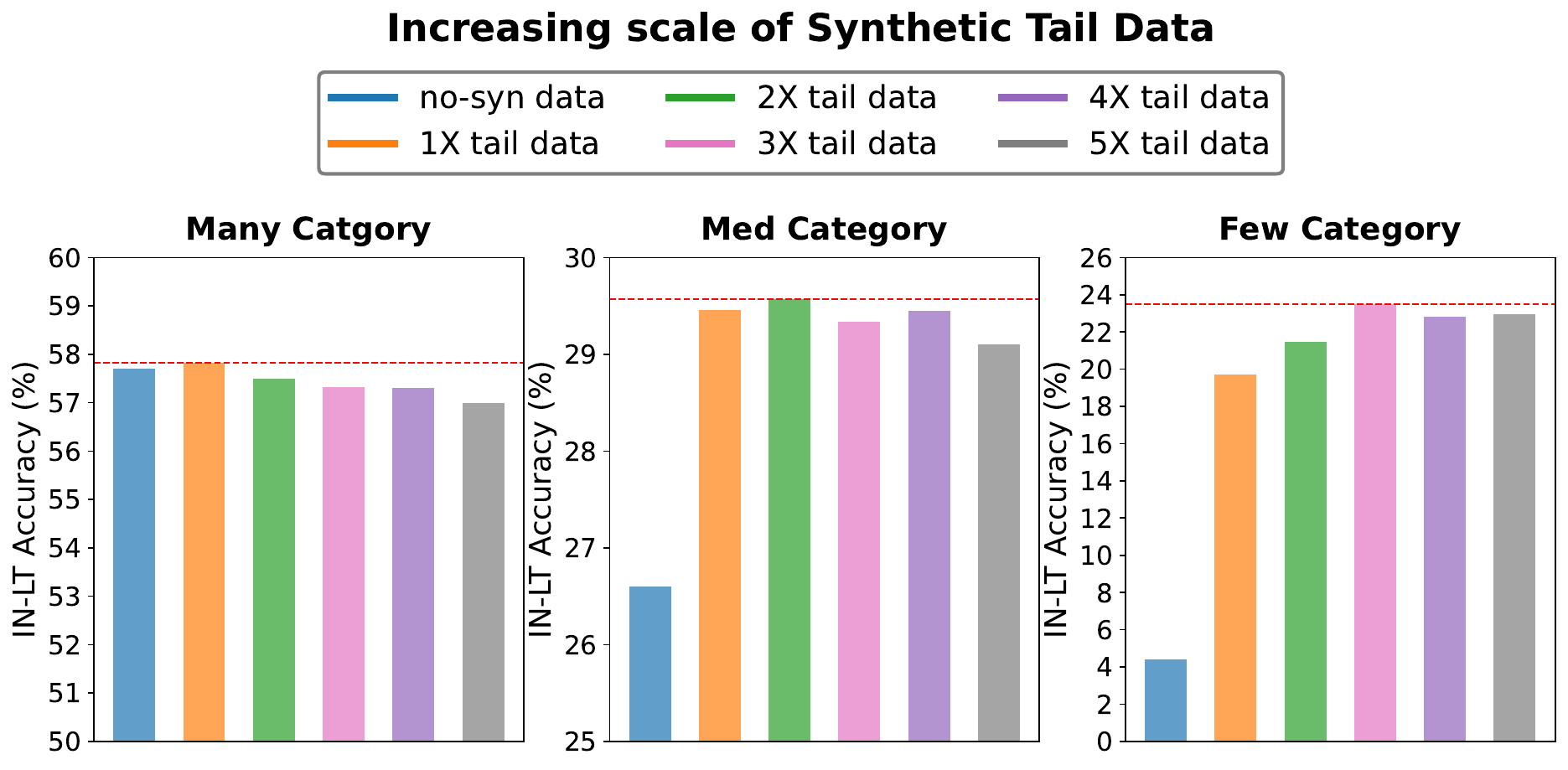}
        \label{fig:wildcurriculum}
    \end{subfigure}
    \vspace*{-1cm}
    \caption{\small{Ablation study on scale of Synthetic data (generated for tail class samples) used in DisCL training with ResNet-10 on IN-LT task. Here $\mathbf{X}$ refers to the original number of tail classes' samples.}}
    \label{fig:scale_r10}
    \vspace*{-1\baselineskip}
\end{figure}

%% file: sections/conclusion.tex
\section{Conclusion}
\label{sec:conclusion}
\vspace*{-0.2cm}

In this paper, we introduce DisCL, a novel paradigm designed to enhance model performance when dealing with low-quality or scarce data. DisCL effectively bridges the distribution gap between original and target data using a spectrum of synthetic data, particularly for challenging samples. Our method utilizes image guidance in diffusion models to generate a comprehensive range of interpolated data from synthetic to real. Additionally, we design specific curricula to maximize the benefits of synthetic data for learning hard samples and closing the gap between synthetic and real data.
The efficacy of DisCL is demonstrated through its significant and robust performance improvements in long-tail classification and learning from low-quality data, across various base model settings. Our analyses reveal that the interpolation of synthetic-to-real data, the selection of guidance intervals, and the proposed curriculum strategy are all essential components contributing to these gains.

Despite the promising results, the performance of DisCL is influenced by certain limitations. The quality of the generated data spectrum is dependent on the capabilities of the diffusion model and the visual-text alignment ability of filtering models. These dependencies constrain the overall performance of DisCL.
Additionally, the current approach to generate text prompts for long-tail classification relies solely on category names derived from large language models (LLMs). To better align with the real data distribution and to reduce the gap between synthetic and real data, future works could focus on generating text prompts from image captions.
Lastly, discrepancies in the position and size of class objects between real and synthetic images can widen the distribution gap. Addressing this issue may involve detecting objects and performing crop operations on real images or using detailed prompts to control these properties in synthetic data. These areas present opportunities for further research and improvement.


%% file: sections/appendix.tex
\section{Appendix}

\vspace*{-0.1cm}
\subsection{Motivation for DisCL's Data Selection}
\vspace*{-0.2cm}
When curating data for a training curriculum, real data often aligns with the test distribution better but suffers from deficiency, noise, low quality, or imbalance; Synthetic data can potentially fix these problems but suffers from a large distribution gap to the test. Our synthetic-to-real curriculum is designed to \textbf{combine the complementary strengths of both data types and overcome their weaknesses}. Unlike previous methods using synthetic data with no real-image guidance or a fixed guidance level, 
DisCL dynamically adjusts the real-image guidance level per training stage to generate a spectrum of synthetic-to-real samples that accelerate learning progress and meanwhile progressively bridging the distribution gap. 
Unlike pre-defined easy-to-hard curricula on real data, DisCL's data selection is adaptive to the training dynamics, considers diversity and distribution gap, and is optimized for achieving the greatest progress per stage. \looseness-1

\vspace*{-0.1cm}
\subsection{Synthetic Data Generation with Image Guidance} \label{sec:app_generation}
\vspace*{-0.2cm}
In this section, we visualize more generated images in (Phase 1) of our method with various levels of image guidance, for two different classification tasks. 

\vspace*{-0.1cm}
\subsubsection{Generation Settings and Statistics} \label{sec:generationsettings}
\vspace*{-0.15cm}
We provide the statistics for the synthetic data generation within our paradigm on ImageNet-LT, CIFAR100-LT, iNaturalist2018, and iWildCam, as shown in Table~\ref{tab:GenerationRatio}.

\input{Table_tex/stats_sync}

\input{Figure_tex/CLIPScore}

\vspace*{-0.1cm}
\subsubsection{ImageNet-LT Synthetic Generation}\label{sec:app_inltgeneration}
\vspace*{-0.1cm}
\paragraph{Selection of Text prompts}\label{sec:app_inltprompt}

To improve model performance on the minority classes, high-quality and diverse synthetic samples are required. To achieve so, we follow the approach in \citet{fu2024dreamda}, and utilize publicly available GPT-3.5-turbo to generate diverse prompts for these 1000 IN-LT classes.
We use the following prompt to query GPT-3.5-turbo for generating descriptions for class \(X\): 

``\textit{Please provide 10 language descriptions for random scenes that contain only the class \(X\) from the ImageNet-LT dataset. Each description should be different and contain a minimum of 15 words. These descriptions will serve as a guide for Stable Diffusion in generating images.}'' 

The sample-prompts generated by GPT-3.5-turbo are listed in Table~\ref{tab:textpromptinlt}.
\input{Table_tex/prompt_inlt}

\vspace*{-0.3cm}
\paragraph{Selection of Images Guidance Levels} \label{sec:app_inltsettings}

We first analyze the cosine similarity between synthetic images and real images, as well as between synthetic images and text prompts. The similarity score between synthetic images and real images can be used to quantify the diversity introduced in the synthetic images. As depicted in Fig.~\ref{fig:imgsimilarity_inlt}, the similarity between synthetic images and real images decrease as the guidance level reduces, demonstrating the trend of increased diversity in the data spectrum. 
However, the changes in the scores are relatively small across varying guidance levels. Combined with the visual cases for this dataset (examples shown in Fig.~\ref{fig:inltsuccessAppend}), we observe that for images generated with high guidance levels (\(\lambda \geq 0.7\)), only minor details are modified by the diffusion model, resulting in high similarity scores above 0.85. However, we aim to provide more diverse synthetic data to increase the model's generalization on the class-balanced test set. Including these highly similar images may hinder the diversity and cause the model to overfit to specific visual features, thereby negatively impacting its generalization ability. Therefore, we select \(\{0.0, 0.1, 0.3, 0.5\}\) as the interval of image guidance levels used in the training process for this dataset.

\vspace*{-0.3cm}
\paragraph{Selection of CLIPScore Threshold}\label{sec:app_inltfilter}
We leverage the widely used CLIPScore \citep{clipscore2022} to filter out poor-quality images in the synthetic data. In this method, the CLIP cosine similarity between synthetic images' embeddings and text embeddings is computed to measure the alignment between images and the corresponding classes provided in text prompts. For the synthetic data generation for ImageNet-LT, we use a unified template that emphasizes the class information in text prompts. Following \citet{trabucco2023effective}, we use "\textit{a photo of <class name>}" to prompt the CLIP model and compute the cosine similarity.
We also consider the value of the filtering threshold for synthetic data. Following previous work \citep{schuhmann2021laion}, we set the threshold to 0.3 based on the distribution of similarity scores and a review of generation quality, as shown in Fig.~\ref{fig:textsimilarity_inlt}. We observe that a threshold of 0.3 effectively filters out synthetic images with poor quality or mismatched classes.

\vspace*{-0.1cm}
\subsubsection{iWildCam Synthetic Generation}\label{sec:app_wildgeneration}

\input{Figure_tex/CLIPScore_iwildcam}

\paragraph{Selection of Text prompts} \label{sec:app_wildprompt}
Following previous work \citep{clark2024text, trabucco2023effective}, we first define prompts for each class using the template "\textit{a photo of <class>}". 
However, the classnames in iWildCam comprises of scientific names, which are usually unseen/unknown concepts to the diffusion text encoder. For example, "canis lupus" is the class name for  "wolf" animal.
To address this, we replace the scientific names with their common names and add a postfix "\textit{in the wild}" in the prompt to drive the generation of wild images. The final text prompt we use is "\textit{a photo of <common name of class> in the wild}".

\vspace*{-0.3cm}
\paragraph{Selection of Images Guidance Levels} \label{sec:app_iwildcamsetting}

Based on the generated data with multiple image guidance scales, we search for effective image guidance scales for this task using CLIP cosine similarity scores between synthetic image embeddings and real image embeddings. As shown in Fig.~\ref{fig:imgsimilarity_wilds}, as the difference between real images and synthetic images increases, the cosine similarity between image embeddings decreases from \(\lambda = 1\) to \(\lambda = 0.3\). However, when the image guidance continues to decrease to \(\lambda = 0\), the cosine similarity score increases slightly.
With low image guidance scales, the diffusion model tends to generate images that heavily rely on text information, maintaining only global information (such as the color of the image background) in the synthetic data for some images. This creates a distribution gap between these synthetic data and real data that is too large for the model to accurately compare the differences between the two images using embedding representation.
Additionally, based on the analysis of the quality of synthetic images and to leverage the difficulty of the features and the distribution gap between synthetic and real data, we set the image guidance scales to \(\{0.5, 0.7, 0.9\}\) for this task.

\vspace*{-0.3cm}
\paragraph{Selection of CLIPScore Threshold}\label{sec:app_wildfilter}
To filter out low-quality images, we assess the CLIP cosine similarity scores between synthetic image embeddings and corresponding text embeddings for each class. 
We use the same prompt template as in the generation process ("\textit{a photo of <common name for animal> in the wild}") to compute CLIPScore for synthetic images. 
The distribution of CLIPScores is shown in Fig.~\ref{fig:textsimilarity_wilds}, which reveals a distinct gap around 0.25. Combined with a review of the quality of synthetic data, we set the threshold to 0.25. Synthetic data with a CLIPScore lower than 0.25 are considered poor-quality samples.

\vspace*{-0.1cm}
\subsubsection{Visualization}\label{sec:app_visualization}

\vspace*{-0.2cm}
\paragraph{Visual Cases}
We provide additional visual examples of synthetic data generated with multiple guidance levels and text prompts for the ImageNet-LT and iWildCam datasets. The results are visualized in Fig.~\ref{fig:inltsuccessAppend} and Fig.~\ref{fig:wildsuccessAppend}. These examples demonstrate that the model can generate synthetic data with various postures, backgrounds, and actions as the image guidance level decreases. Particularly for ImageNet-LT generation results, diverse prompts introduce more varied features into low-guidance data. These diverse features enable the model to achieve better generalization on the target distribution.

\input{Figure_tex/inlt_sample1}
\input{Figure_tex/iwildcam_sample1}

\vspace*{-0.3cm}
\paragraph{Failure Cases}\label{sec:app_failure}
During generation, despite designing text prompts and applying CLIPScore to filter to remove low-quality data, some failure cases still occur in the synthetic dataset. In this section, we discuss these failure cases encountered during the generation process.
As shown in Fig.~\ref{fig:inltfailureAppend} and Fig.~\ref{fig:wildfailureAppend}, the first failure case is caused due to the inability to recognize objects in the original images. If these objects are  clearly obscured or hard-to-identify (e.g. second case in Fig.~\ref{fig:wildfailureAppend} and first case in Fig.~\ref{fig:inltfailureAppend}), diffusion models cannot accurately identify the object or modify details for generating diverse and useful data. For these seed images, only synthetic data generated with a low-guidance scale can achieve a CLIPScore higher than the threshold. However, this approach compromises the smooth transition of data from synthetic to real distribution.
Even though the diffusion model can generate images with a smooth transition for most-of-the-cases, our quality-check on synthetic data can constrain the feature extraction and alignment ability of the CLIP model. For example, in second case of Fig.~\ref{fig:inltfailureAppend},  CLIPScore  filters out the slightly modified but perceptually useful images, containing prototypical class features.

\input{Figure_tex/inlt_failure}
\input{Figure_tex/iwild_failure}

\subsection{Application of DisCL to Other Datasets and Model Scale}
\label{sec:app_otherdataset}
\vspace*{-0.2cm}

To further assess the robustness of DisCL, we extend our experiments to two additional widely used imbalanced datasets: CIFAR-100-LT \citep{cao2019learning} and iNaturalist2018 \citep{van2018inaturalist}. 
For iNaturalist2018, we generate synthetic data following the same approach and settings used for the long-tail classification task on ImageNet-LT. In the case of CIFAR-100-LT dataset, due to the lower image resolution, we adjust the image guidance scale to $\{0.5, 0.7, 0.9\}$ so as to ensure high-quality synthetic data generation. Visual examples of the generated data are shown in Fig.~\ref{fig:cifar} and ~\ref{fig:inat18}. 
For CIFAR-100-LT, we evaluate the performance of DisCL under different imbalance ratios (50 and 100).
Additionally, we expand our model evaluation to a larger scale, ResNet-34 (widely adopted for ImageNet) with the same experimental settings of DisCL as before. 
As evident from Table~\ref{tab:cifar100} and Table~\ref{tab:inat2018}, our results demonstrate that DisCL achieves a notable  improvements in overall top-1 accuracy (e.g., +1–3.3$\%$ over baselines) and few class performance (e.g., +3-8$\%$ for tail classes) across both datasets. We also notice that combining a class-reweighting loss (BS) with DisCL causes an oversaturation in tail-class signals, causing the model to neglect \textit{many} classes during the training. This suggests that reweighting and mixing synthetic with real data address different aspects of class imbalance; aligning with prior works, \cite{ahn2023cuda} and \cite{resampleLT2023}.
\textit{Notably}, the top-1 accuracy gains persist when scaling the model to ResNet-34, as demonstrated for CIFAR-100-LT in Table~\ref{tab:cifar_resnet34} and ImageNet-LT in Table~\ref{tab:imagenet_resnet34}. This underscores the flexibility of our proposed DisCL method across different datasets and model scales.

\input{Table_tex/resnet34_cifar_inat_lt}

\input{Figure_tex/cifar_samples}

\input{Figure_tex/inat_samples}

\vspace*{-0.1cm}
\subsection{Training with Curriculum Learning}\label{sec:app_trainingimplement}

\vspace*{-0.1cm}
\subsubsection{Long-Tail Learning with Non-Adaptive Strategy}
\label{sec:app_inltimplementation}
\vspace*{-0.1cm}
For long-tail classification, we propose a non-adaptive curriculum learning strategy that starts with the lowest guidance and progressively increases to the highest guidance within the defined interval \(\Lambda\). 
We employ a linear scheduler to adjust the guidance levels during training, allowing the model to train with data from various guidance levels for \textit{equal durations}. 
Furthermore, the test set of ImageNet-LT is in-distribution to its training data; unlike the training data, it is a class-balanced set. To mitigate the potential negative effects of the distribution gap between synthetic and real data, all the hard tail samples from original data are involved into training at all times. 
Furthermore, with DisCL, number of samples for tail classes increases along with the introduction of synthetic data at each  stage, however the ratio of tail-to-nontail samples is still very skewed. To preserve a constant imbalance-ratio throughout all training stages and experiments, we undersample the non-tail samples at "each stage" so that ratio of tail-samples to non-tail samples matches the proportion of tail classes to non-tail classes present in the original data (13.6$\%$).

All experiments are conducted based on this proportion setting. Complete strategy details are covered in Algorithm~\ref{alg:DisCLnonadaptive}.

\vspace*{-0.1cm}

\input{Algorithm_tex/nonadaptive}

\vspace*{-0.1cm}
\subsubsection{Learning from Low-Quality Data with ``Adaptive Curriculum'' Strategy}
\label{sec:app_wildimplementation}
\vspace*{-0.25cm}
An approximation method to assess the effectiveness of samples in helping model achieve greatest progress on and fastest learning face is introduced by DoCL \citep{zhou2021curriculum} as shown in Eq~\ref{eq:progressassessment}. 
\begin{equation}
\label{eq:progressassessment}
    \begin{split}
        \mathbb{E}_{x\in D, x\sim \mathcal{D}} \langle y - f(x), \frac{\partial f(x)}{\partial t} |_S\rangle \\
        \approx \frac{1}{|D|}\sum_{j\in \mathcal{V}} \langle y^{(j)} - f(x^{(j)}), \frac{\partial f(x^{(j)})}{\partial t} |_D \rangle
    \end{split}
\end{equation}
where \(\mathcal{D}\) is the training distribution and \(x\in D\) is a set of finite samples randomly sampled from the original distribution \(\mathcal{D}\). \(\mathcal{V}\) denotes the subset of samples from \(S\). Here, \(y\) and \(f(x)\) denotes the target-class and sample prediction. \(\langle y-f(x), \frac{\partial f(x)}{\partial t} |_\mathcal{V}\rangle\) represents the project of residual \(y-f(x)\) on the model dynamics \(\frac{\partial f(x)}{\partial t} |_\mathcal{V}\). 
This equation indicates that when trained with subset \(\mathcal{V}\), the expected progress \(\mathbb{E}\) of samples in the original training dataset can be approximated by the progress of samples on subset \(\mathcal{V}\) achieved via training on the set \(D\). 

For learning from low-quality data, we adopt DoCL and implement an adaptive curriculum strategy to select the synthetic data with best guidance level for each training stage. We showcase the implementation in Algorithm \ref{alg:DisCLadaptive}, wherein we preserve $i$ for indexing the guidance level in \(\Lambda\) and $j$ for indexing the sample in a given dataset.
Before the training process, we randomly select samples from the spectrum for each guidance level in \(\Lambda\) and mark it as guidance validation set \(\mathcal{V}\) for progress evaluation. This set has zero overlap with the training data $\mathcal{D}_{\text{all}}$.
At each training stage, we randomly sample a set \(D\) (termed as random-real set) from the training dataset $\mathcal{D}_{\text{all}}$. Before selecting the guidance level, we train the model on dataset \(D\) and evaluate the progress (in terms of classifier's prediction score) achieved on samples of each subset \(\mathcal{V}_{i}\) corresponing to a given guidance $\lambda_i$. We then select the $\lambda_i$ with the highest progress to gather synthetic data and combine it  with other non-hard samples from the original training data for the current training stage. This technique encourages the model to adaptively select the most informative guidance for the current training stage. At the end of the curriculum-training, to alleviate the negative effect of the distribution gap between synthetic data and real data for this task, we keep finetuning the model with real data for a short period. 
The steps of algorithm are detailed in Algorithm~\ref{alg:DisCLadaptive}.

\vspace*{-0.1cm}

\input{Algorithm_tex/adaptive}

\vspace*{-0.1cm}

\subsection{Hyperparameters for Synthetic Generation and Model Training} \label{sec:app_hyper}
\vspace*{-0.2cm}
The values of all hyperparameters used for synthetic data generation with diffusion model and curriculum learning strategy are listed in Table~\ref{tab:hyperparameter}.

For ImageNet-LT, we implement baselines based on the codebase and the pretrained model from \href{https://github.com/AlvinHan123/LDMLR/tree/main}{LDMLR}. We also re-implement CUDA baseline from this \href{https://github.com/sumyeongahn/CUDA_LTR/tree/main}{codebase}, containing some missing models. We use the same hyper-parameter settings as listed in the CUDA paper. For FLYP, we implement baseline models with \href{https://github.com/locuslab/FLYP}{FLYP codebase} and leverage the available pretrained model from \href{https://github.com/mlfoundations/open_clip}{Open CLIP}.

\vspace*{-0.1cm}
\subsection{Computational Requirements for Synthetic Generation}
\vspace*{-0.2cm}
For computational requirements of offline generation, $1$ RTX A5000 GPU is used to generate synthetic images. For time efficiency, It took ~$10$ seconds to generate a full spectrum ($6$ image guidance levels) of synthetic images for each real image with resolution=$480 \times 270$. 

\input{Table_tex/hyperparameters}

\vspace*{-0.1cm}
\subsection{Further Discussion on Experiment Results}
\vspace*{-0.2cm}

In this section, we analyze the results of each guidance level under \textit{Fixed Guidance} experiment to observe the effect of different image guidance levels on the classifier's performance. During the training process, synthetic data generated from only a specific guidance level combined with original real data is presented to the model. The ablation numbers are shown in Fig.~\ref{fig:GuidanceDifference}. 

For the iWildCam dataset, data generated with text-only guidance (\( \lambda = 0 \)) has the largest distribution gap between synthetic and real data, and it also showcases lowest Out-of-Distribution (OOD) performance. As the guidance scale increases, this distribution gap diminishes, and the OOD F1 score consistently improves. This outcome aligns with the visually observed reduction in distribution differences between generated and real images. 

Conversely, the trend seen with ImageNet-LT  diverges from above. In long-tail classification, we aim to increase data diversity while keeping the distribution gap small. As detailed in Appendix~\ref{sec:app_inltsettings}, on one hand, generating synthetic data that closely resemble real data further reduces the diversity, and generating synthetic data far from real distribution can offer diversity but hurt OOD performance. In case of ImageNet-LT, we observe that more diverse synthetic data tends to significantly improve the classifiers' generalization.

Inspired by these observations, we tailor our guidance scales intervals according to the task-at-hand.

\input{Figure_tex/ablation_guid}

\vspace*{-0.1cm}
\subsection{Improvement on Worst-$k$ classes: Balanced Softmax (BS)  v/s DisCL with BS \cite{bs2020}}
\vspace*{-0.2cm}

While DisCL's average gain over Balanced Softmax Baseline(BS) is $+2.07\%$, it improves BS's worst-$k$ class accuracy by $4.5\%$–$7.6\%$, verifying our targeted advantage on the most difficult classes—precisely where strong baselines struggle.
It demonstrates that DisCL complements existing methods, improving performance where it matters most, even compared with strong baselines.
\input{Table_tex/compare_bs}


\vspace*{-0.1cm}
\subsection{Societal Impact}
\label{sec:society}
\vspace*{-0.2cm}

Our proposed method is beneficial for diverse fields, where inadequate quantity and low quality of data is common, \textit{e.g.} medical domain. The synthetic data generation, as followed by DisCL approach can reduce the need for extensive data collection, therefore mitigating the ethical concerns related to data-privacy. Overall, our method DisCL can democratize the access of effectively training ML models  in the low-resource environments. However, by leveraging the pretrained generative models, the potential biases of models can perpetuate into the synthetic data and eventually affect the sensitive real-world applications consuming this data, such as medical diagnosis, law enforcement \textit{etc.}


\clearpage

%% file: Table_tex/stats_sync.tex
\begin{table*}[ht]
\centering
\resizebox{0.8\linewidth}{!}{%
\begin{tabular}{@{}l|c|cc|c|c@{}}
\toprule
\textbf{Images' Details }     & \textbf{ImageNet-LT} & \multicolumn{2}{c|}{\textbf{CIFAR100-LT}} & \textbf{iNaturalist2018} & \textbf{iWildCam} \\
 & & \small{\textbf{Irb=100}} & \small{\textbf{Irb=50}} & & \\
\midrule
No. of Hard Samples  & 1643 & 324 & 268 & 44956 & 8260  \\
Number of Image Guidance Scales \(\lambda\) & 4 & 4 & 4 & 4 & 3 \\
Number of Random Seed Per Image     & 8 & 8 & 8 & 4 & 8   \\ 
Number of Generated Images      & 51917 & 2592 & 2144 & 179824 & 197756   \\ 
Number of Generated Images After Filtering   & 24141  & 809 & 668 & 75234 & 90093  \\ 
Acceptance Rate   & 46.50\%  & 31.21\% & 31.16\% & 41.84\% & 45.56\%  \\ 
\bottomrule
\end{tabular}
}
\vspace*{0.2cm}
\caption{Statistics about Generated Synthetic Data. Irb refers to the imbalance ratio used to sample CIFAR100-LT dataset.}
\label{tab:GenerationRatio}
\vspace*{-0.2cm}
\end{table*}

%% file: Figure_tex/CLIPScore.tex
\begin{figure*}[t]
    \centering
    \begin{subfigure}{0.45\linewidth}
        \centering
        \includegraphics[width=\linewidth]{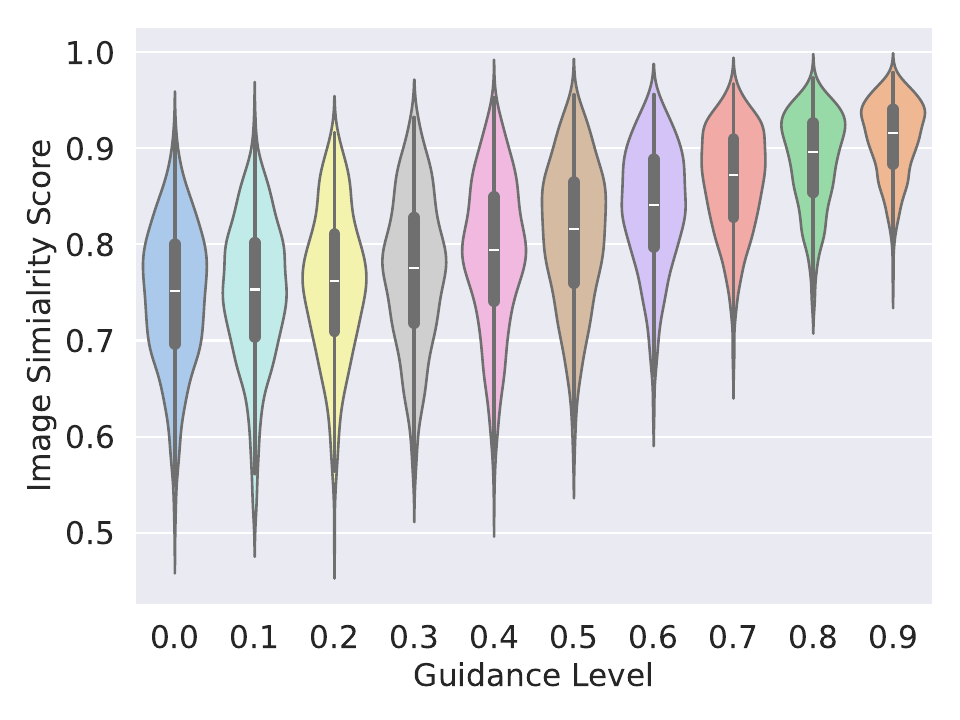}
        \caption{Similarity b/w synthetic images \(\&\) its original real image. }
        \label{fig:imgsimilarity_inlt}
    \end{subfigure}
    \hfill
    \begin{subfigure}{0.45\linewidth}
        \centering
        \includegraphics[width=\linewidth]{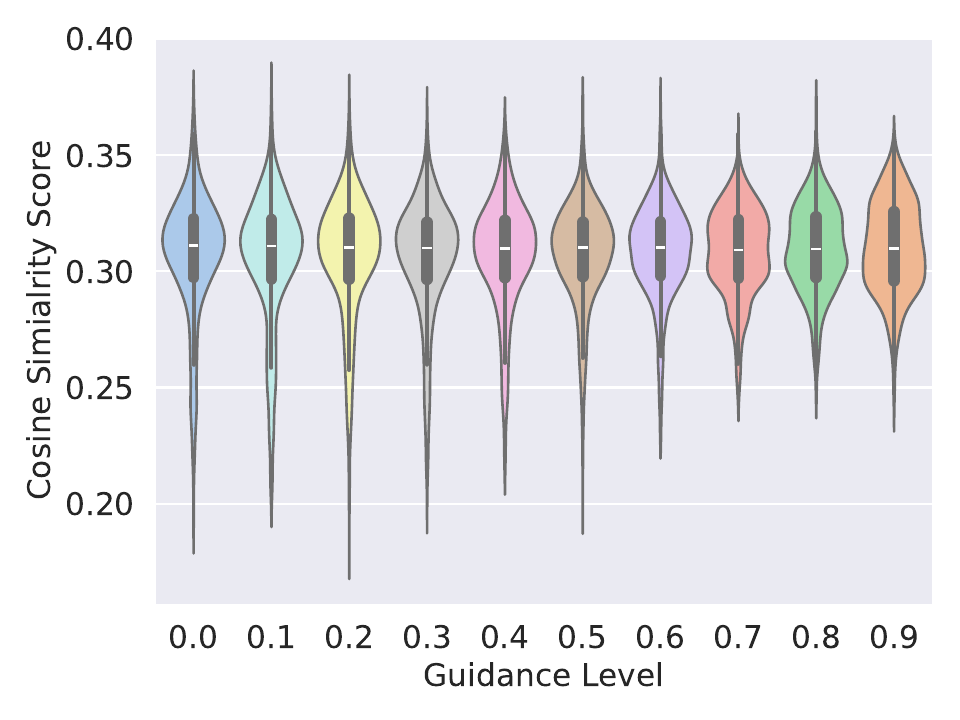}
        \caption{Similarity b/w synthetic images \(\&\) defined text prompt.}
        \label{fig:textsimilarity_inlt}
    \end{subfigure}
    \caption{CLIP Cosine similarity score for ImageNet-LT Synthesis.}
    \vspace*{-1\baselineskip}
\end{figure*}

%% file: Table_tex/prompt_inlt.tex
\begin{table*}[ht]
\centering
\begin{tabular}{c| p{0.7\linewidth}}
\toprule
\textbf{Class Name}     & \textbf{Prompts}  \\
\midrule
Grand Piano & A grand piano sits elegantly in a sunlit room, its glossy finish reflecting the warm glow.  \\
 & In a cozy living room, the grand piano adds a touch of luxury and sophistication to the space. \\
 & The grand piano sits silently in a dimly lit room, waiting patiently for a skillful pianist to bring it to life.   \\ 
 & In a grand ballroom, the grand piano provides a majestic backdrop for a glamorous event.   \\ 
 & A vintage grand piano exudes timeless elegance in a quaint parlor, filled with antique charm.  \\ 
\midrule
Pufferfish & A colorful pufferfish swimming gracefully in a crystal-clear ocean, surrounded by vibrant coral reefs.  \\
 & A group of playful pufferfish blowing bubbles and chasing each other in a sunlit underwater cave. \\
 & A shoal of pufferfish moving in unison, creating a mesmerizing dance of synchronized swimming in the deep sea.   \\ 
 & A fierce pufferfish defending its territory from intruders, puffing up its body and displaying its sharp spikes as a warning.   \\ 
 & A baby pufferfish following its larger parent closely, learning the ropes of survival in the vast ocean ecosystem. \\ 
\bottomrule
\end{tabular}
\vspace*{-0.2cm}
\caption{Generated text prompts for ImageNet-LT classes}
\vspace*{0.2cm}
\label{tab:textpromptinlt}
\vspace*{-1\baselineskip}
\end{table*}

%% file: Figure_tex/CLIPScore_iwildcam.tex
\begin{figure*}[t]
    \centering
    \begin{subfigure}{0.45\linewidth}
        \centering
        \includegraphics[width=\linewidth]{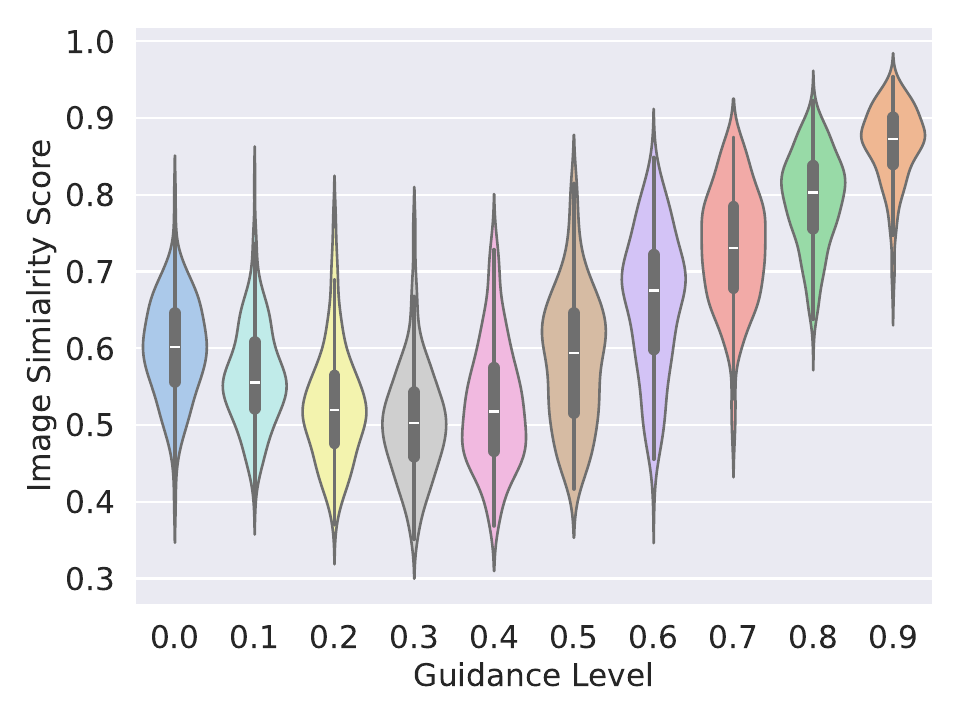}
        \caption{Synthetic image \(\&\) original real images. }
        \label{fig:imgsimilarity_wilds}
    \end{subfigure}
    \hfill
    \begin{subfigure}{0.45\linewidth}
        \centering
        \includegraphics[width=\linewidth]{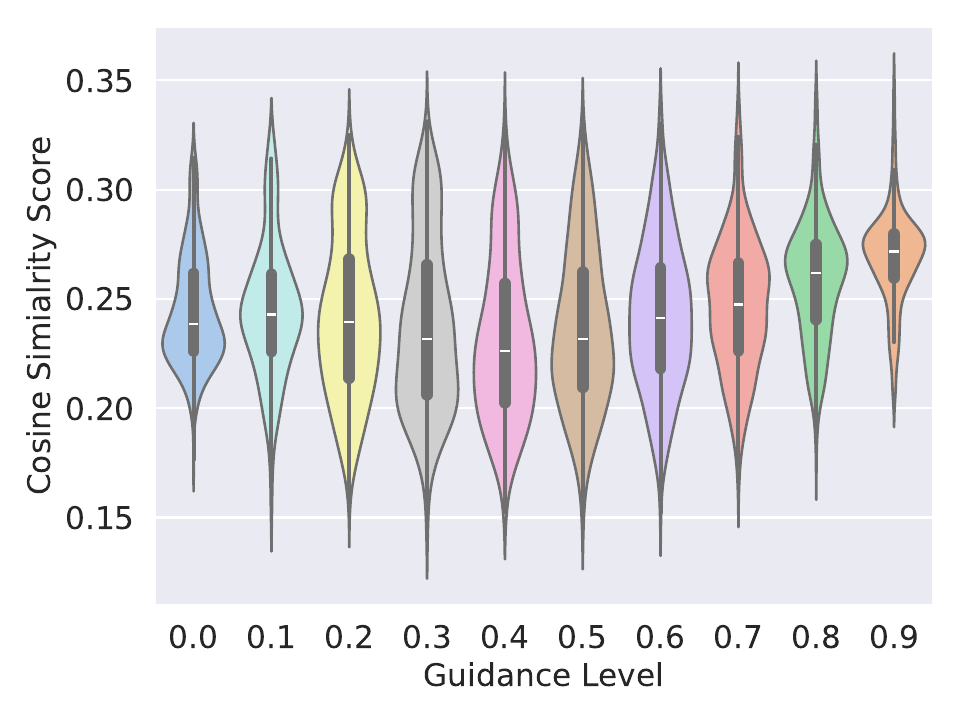}
        \caption{Synthetic image \(\&\) defined text prompt.}
        \label{fig:textsimilarity_wilds}
    \end{subfigure}
    \caption{CLIP Cosine similarity score for iWildCam Synthesis. }
    \vspace*{-1\baselineskip}
\end{figure*}

%% file: Figure_tex/inlt_sample1.tex
\begin{figure*}[ht]
    \centering
    \includegraphics[width=0.7\textwidth]{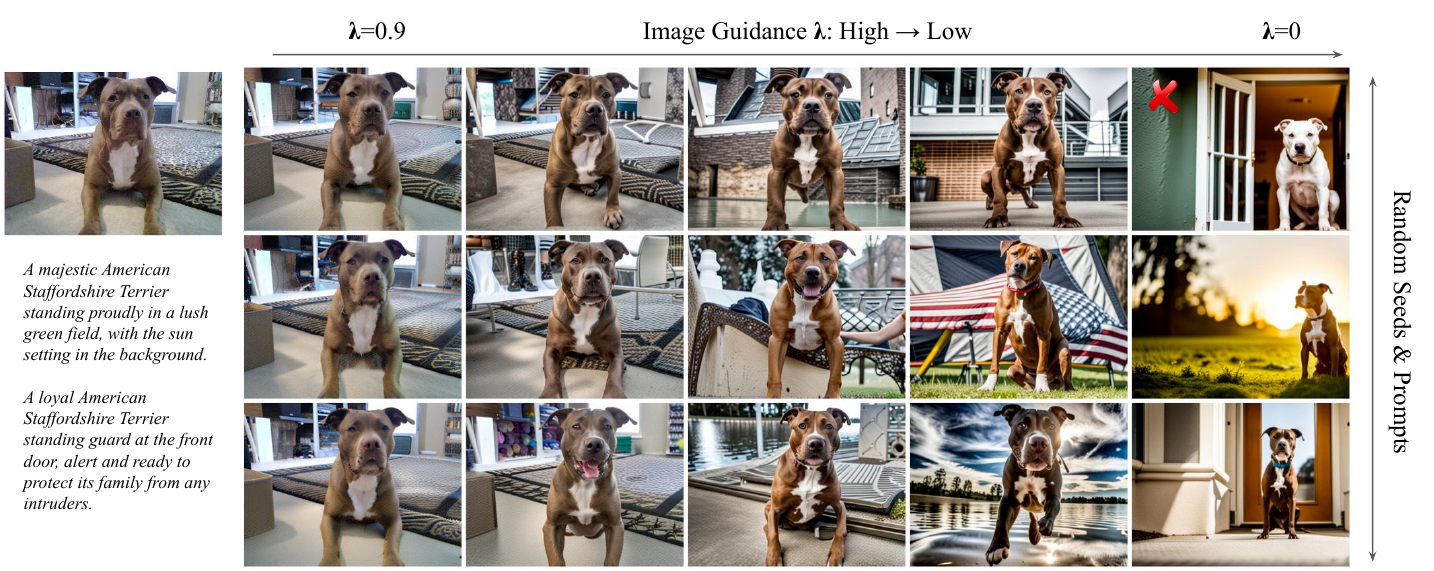}
    \includegraphics[width=0.7\textwidth]{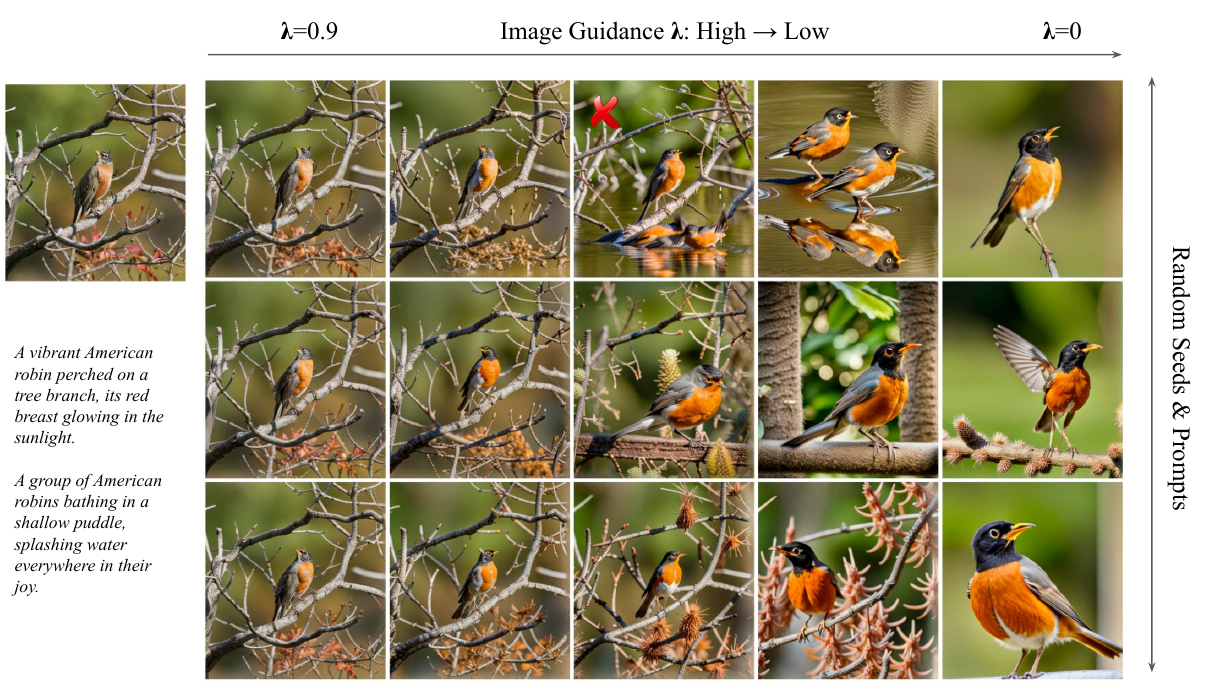}
    \includegraphics[width=0.7\textwidth]{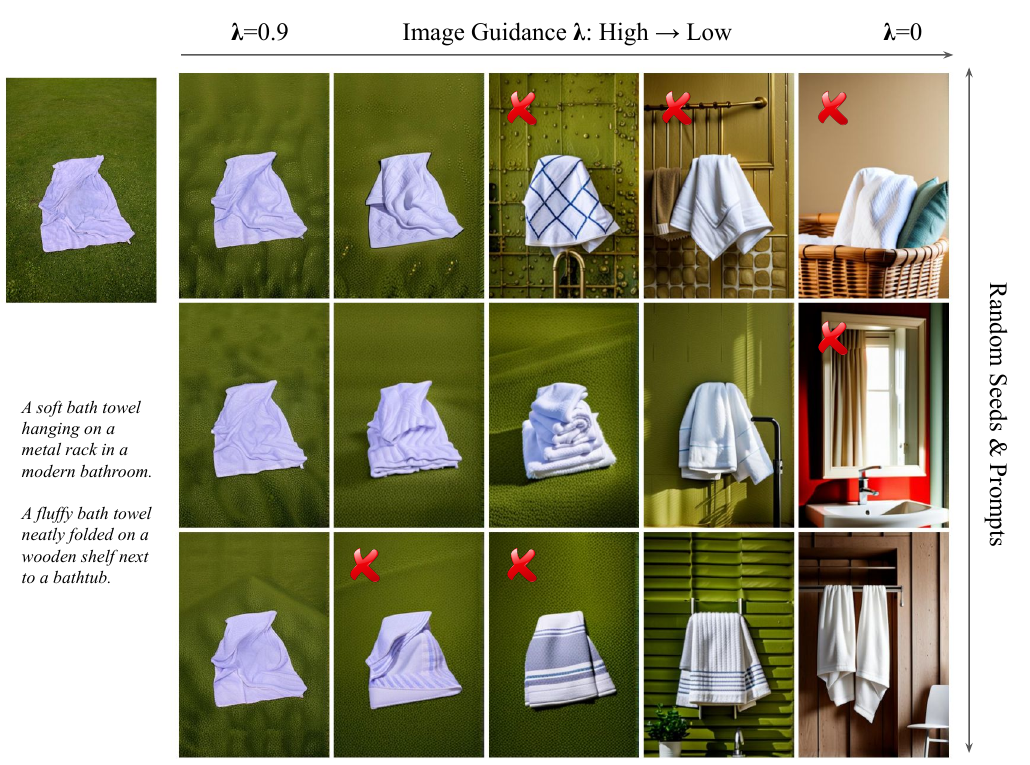}
    \caption{Synthetic generation with various image guidance and random seeds based on ImageNet-LT.}
    \label{fig:inltsuccessAppend}
    \vspace*{-1\baselineskip}
\end{figure*}

%% file: Figure_tex/iwildcam_sample1.tex
\begin{figure*}[ht]
    \centering  
    \includegraphics[width=0.9\textwidth]{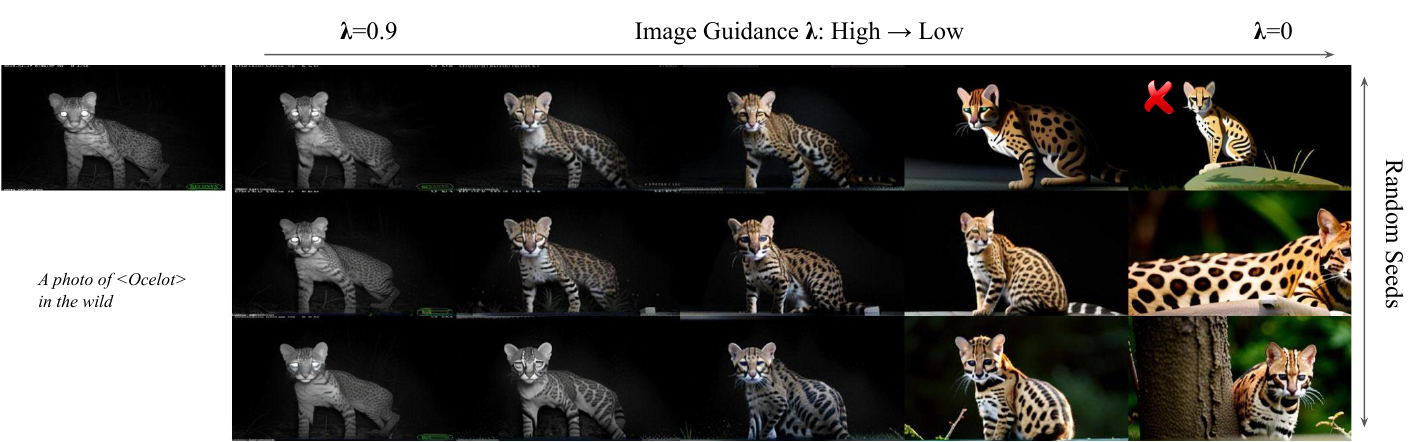}
    \includegraphics[width=0.9\textwidth]{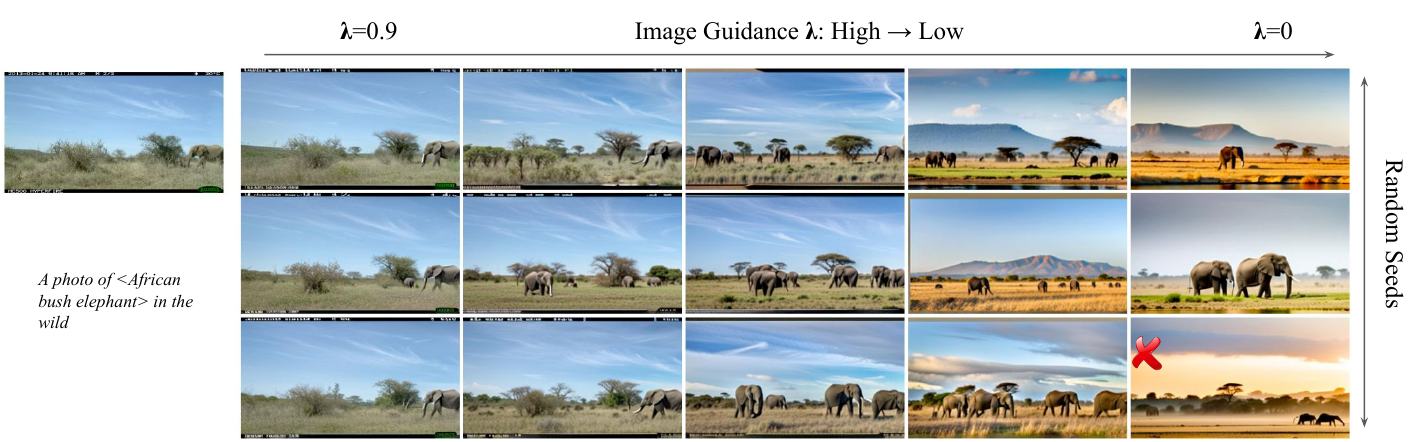}
    \includegraphics[width=0.9\textwidth]{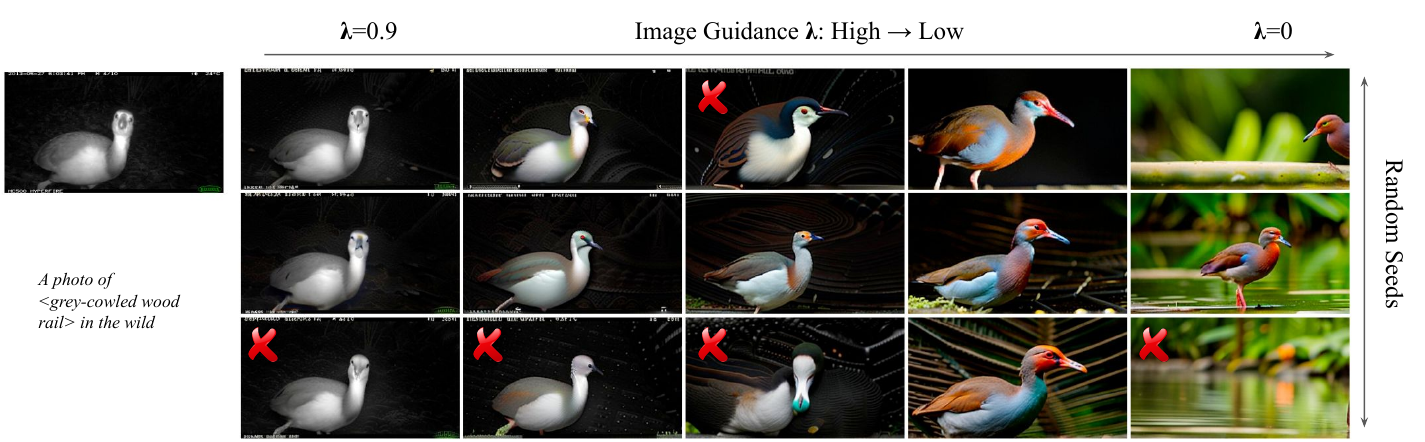}
    \caption{Synthetic generation with various image guidance and random seeds based on iWildCam.}
    \label{fig:wildsuccessAppend}
    \vspace*{-0.5\baselineskip}
\end{figure*}

%% file: Figure_tex/inlt_failure.tex
\begin{figure*}[ht]
    \centering
    \includegraphics[width=0.9\textwidth]{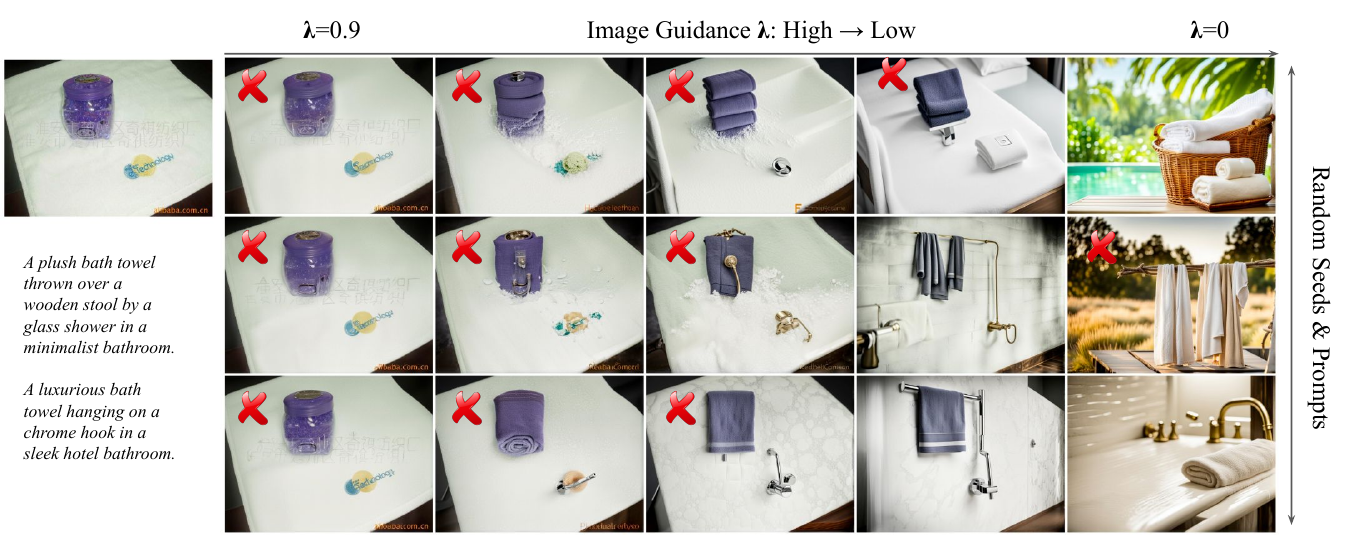}
    \includegraphics[width=0.9\textwidth]{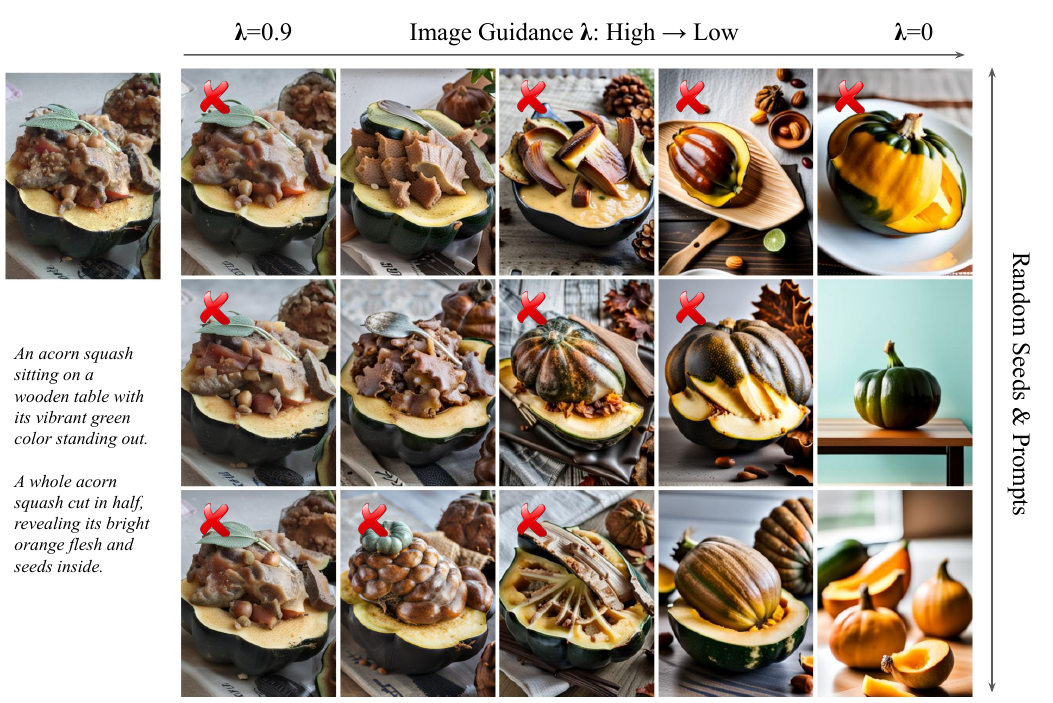}
    \includegraphics[width=0.9\textwidth]{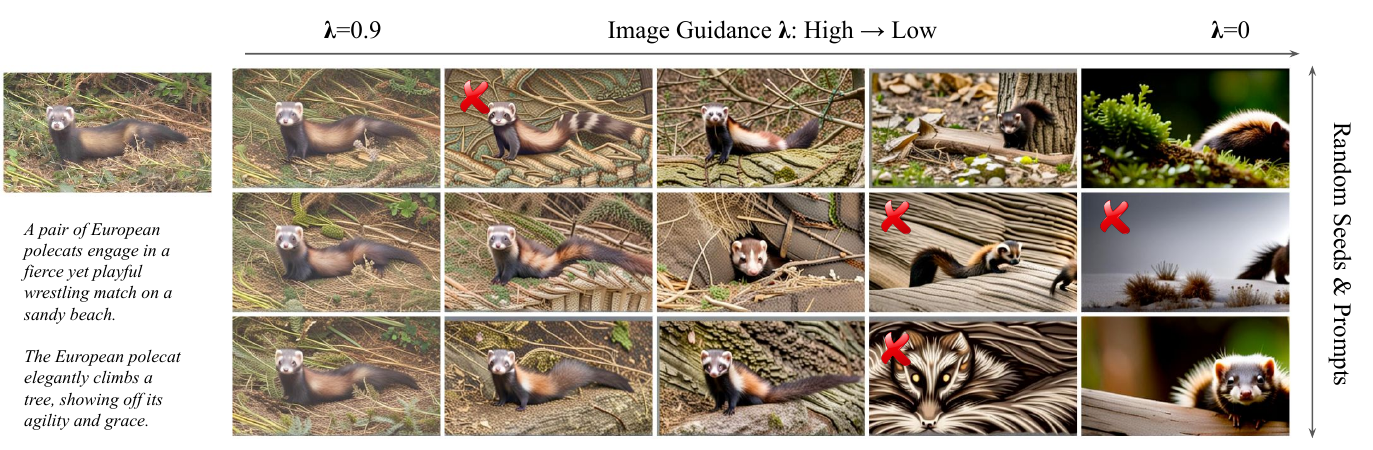}
    \caption{Failure cases for ImageNet-LT synthetic generation}
    \label{fig:inltfailureAppend}
    \vspace*{-1\baselineskip}
\end{figure*}

%% file: Figure_tex/iwild_failure.tex
\begin{figure*}[ht]
    \centering
    \includegraphics[width=0.9\textwidth]{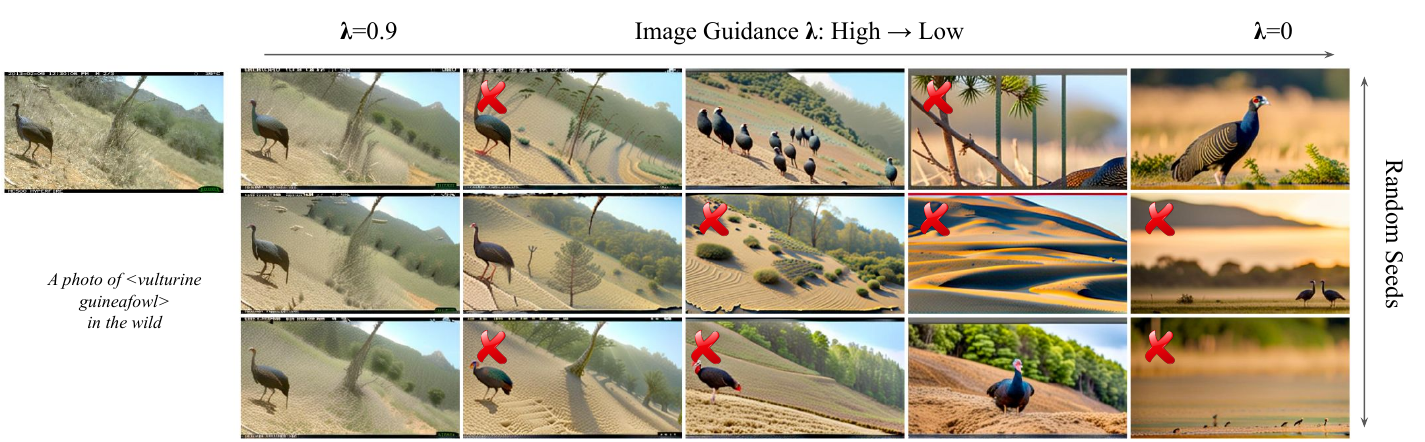}
    \includegraphics[width=0.9\textwidth]{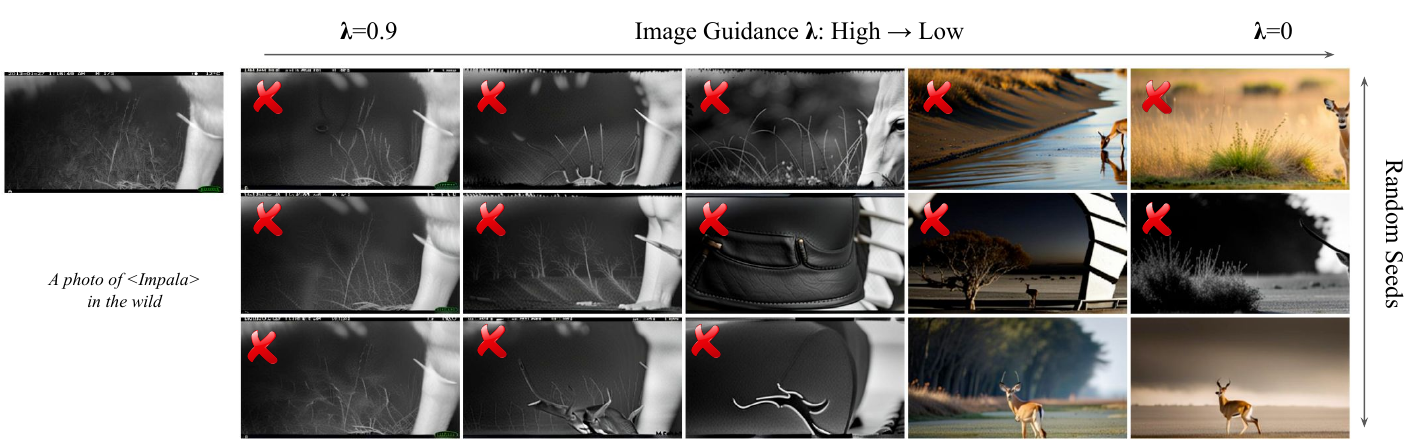}
    \includegraphics[width=0.9\textwidth]{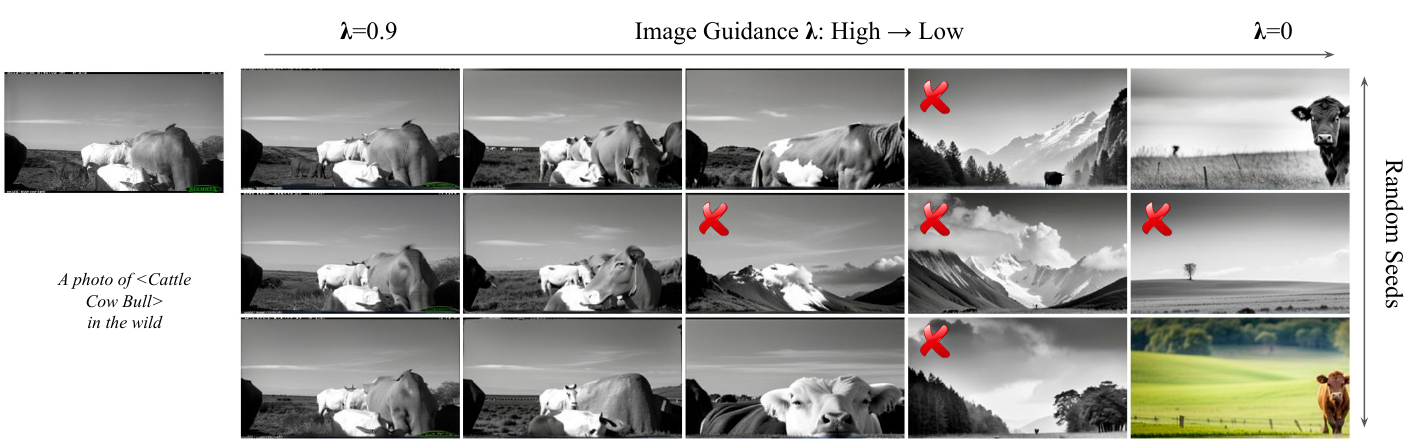}
    \caption{Failure cases for iWildCam synthetic generation}
    \label{fig:wildfailureAppend}
    \vspace*{-0.5\baselineskip}
\end{figure*}

%% file: Table_tex/resnet34_cifar_inat_lt.tex
\begin{table*}[t]
\centering
\resizebox{0.85\linewidth}{!}{%
\begin{tabular}{lc|cccc|cccc}
\toprule
& & \multicolumn{4}{c|}{\textbf{CIFAR-100-LT (Imbalance Ratio=100)}} & \multicolumn{4}{c}{\textbf{CIFAR-100-LT (Imbalance Ratio=50)}} \\
\textbf{Method} & \textbf{Curriculum} & \textbf{Many} & \textbf{Medium} & \textbf{Few} & \textbf{Overall} & \textbf{Many} & \textbf{Medium} & \textbf{Few} & \textbf{Overall} \\
\midrule
\small{CE} &N/A & \textbf{51.71} & 23.51 & 5.05 & 27.7 & \textbf{52.14} & \textbf{29.97} & 10.7 & 32.04 \\
\textbf{\small{CE + DisCL}} & Diverse to Specific & 49.83  & 23.26 & \textbf{7.9} & \textbf{28.4} & 51.83 & 29.12 & \textbf{12.64} & \textbf{32.18} \\
\midrule
\small{\textcolor{orange}{BS}} &N/A & \textbf{46.23} & \textbf{28.0} & 13.13 & 29.79 & \textbf{46.48} & \textbf{33.48} & 22.1 & 34.6 \\
\textbf{\small{\textcolor{orange}{BS} + DisCL}} & Diverse to Specific & 44.9 & 27.4 & \textbf{16.8 }& \textbf{30.3}& 45.51 & 32.08 & \textbf{23.99} & 34.5\\
\bottomrule
\end{tabular}
}
\vspace*{-0.2cm}
\caption{
\small{Accuracy (\%) of ResNet-34 on CIFAR-100-LT classification task with imbalance ratios of 100 and 50, highlighting the best accuracy in bold for overall and class categories (\emph{many}, \emph{medium}, and \emph{few}).}}
\vspace*{-1\baselineskip}
\label{tab:cifar_resnet34}
\end{table*}
\begin{table*}[t]
\centering
\resizebox{0.55\linewidth}{!}{%
\begin{tabular}{lc|cccc}
\toprule
& & \multicolumn{4}{c}{\textbf{ImageNet-LT}} \\
\textbf{Method} & \textbf{Curriculum} & \textbf{Many} & \textbf{Medium} & \textbf{Few} & \textbf{Overall} \\
\midrule
\small{CE}                 & N/A                     & 63.01    & 35.90    & 10.10    & 42.98    \\ 
\small{CE + CUDA}                 & N/A                     &  62.78 & 36.91 & 11.92 & 43.34   \\ 
\textbf{\small{CE + DisCL}} & Diverse to Specific & \textbf{63.54} & \textbf{36.93} & \textbf{13.64} &  \textbf{44.26} \\
\midrule
\small{\textcolor{orange}{BS}} &N/A & \textbf{62.78}	& 36.91 & 	11.92	& 43.34 \\
\small{\textcolor{orange}{BS} + CUDA}                 & N/A                     &  57.16 & 44.5 & 30.49 & 47.33 \\ 
\textbf{\small{\textcolor{orange}{BS} + DisCL}} & Diverse to Specific & 58.82	& \textbf{45.21} & \textbf{32.53}& \textbf{48.42}\\
\bottomrule
\end{tabular}%
}
\vspace*{-0.1cm}
\caption{
\small{Accuracy (\%) of ResNet-34 on ImageNet-LT classification task, highlighting the best accuracy in bold for overall and class categories (\emph{many}, \emph{medium}, and \emph{few}).}}
\vspace*{-1\baselineskip}
\label{tab:imagenet_resnet34}
\end{table*}

%% file: Figure_tex/cifar_samples.tex
\begin{figure*}[ht]
    \centering  
    \includegraphics[width=\columnwidth]{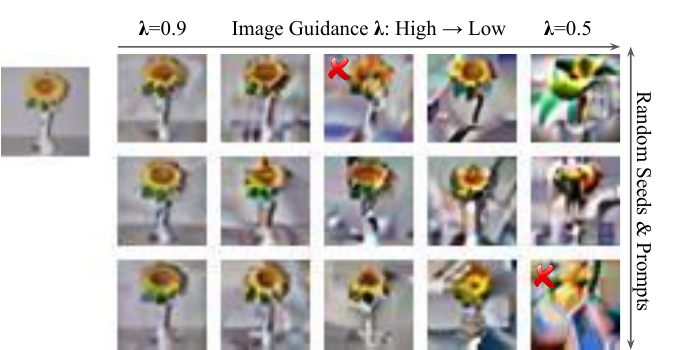}
    \hfill
    \includegraphics[width=\columnwidth]{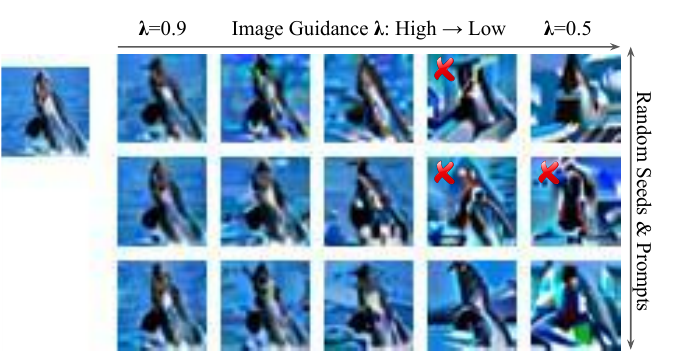}
    \caption{Synthetic generation with various image guidance and random seeds based on CIFAR100. Sample Prompt: \textit{(1) A bright sunflower standing tall in a field, basking in the warm sunlight of a summer day.} \textit{(2) A majestic whale breaches the surface of the deep blue ocean, sending a spray of water into the air.} }
    \label{fig:cifar}
    \vspace*{-1\baselineskip}
\end{figure*}

%% file: Figure_tex/inat_samples.tex
\begin{figure*}[ht]
    \centering  
    \includegraphics[width=0.85\linewidth]{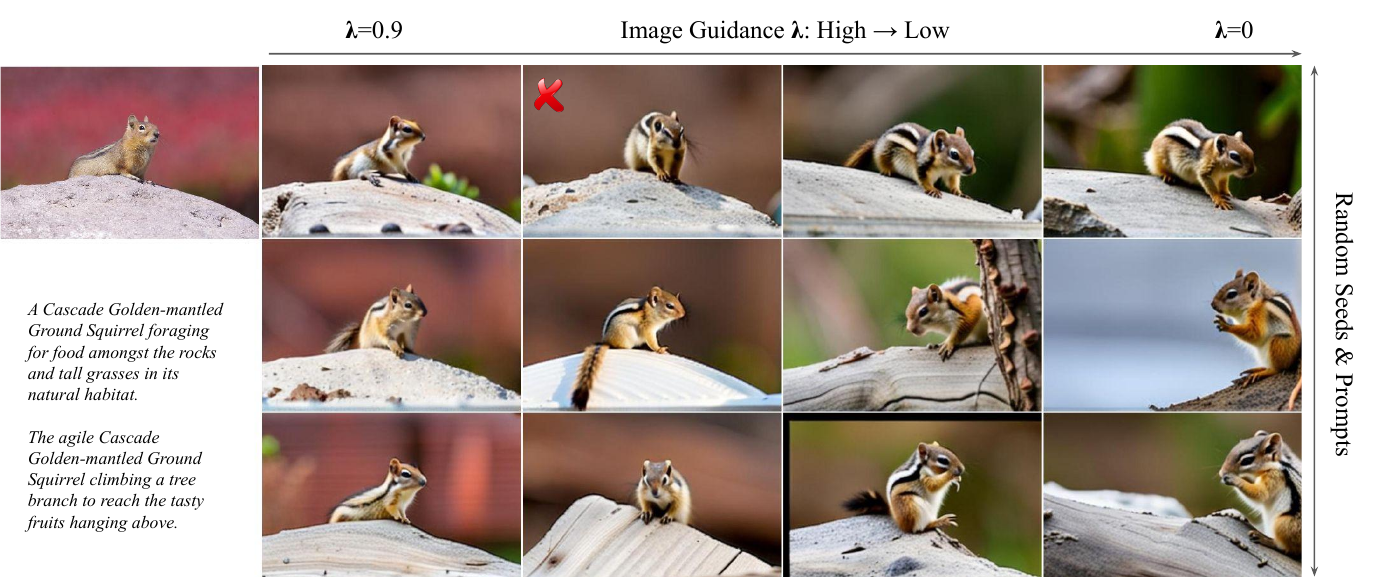}
    \caption{Synthetic generation with various image guidance and random seeds based on iNaturalist 2018.}
    \label{fig:inat18}
    \vspace*{-1\baselineskip}
\end{figure*}

%% file: Algorithm_tex/nonadaptive.tex
\begin{algorithm}[b]
    \SetKwInput{KwInput}{Input}
    \SetKwInput{kwInitialize}{Initialize}
    \SetKwInput{KwOutput}{Output}
    
    \KwInput{
    Image guidance levels \(\Lambda = \{\lambda_i \mid \lambda_i \in [0, 1]\}\), \\
    Non-hard samples \(\mathcal{D}_{\text{nh}} = \{(x^{(j)}, y^{(j)}, \lambda^{(j)} = 1)\}_{j=1}^N\), \\
    Spectrum of syn-to-real data \(\mathcal{S} = \{(x'^{(j)}, y^{(j)}, \lambda^{(j)}) \mid \lambda^{(j)} \in \Lambda\}_{j=1}^M\), \\
    Original hard samples \(\mathcal{D}_{\text{h}} = \{(x^{(j)}, y^{(j)}, \lambda^{(j)} = 1) \mid (x^{(j)}, y^{(j)}, \lambda^{(j)}) \in \mathcal{S}\}\), \\
    Total training epochs \(E\), curriculum cutoff \(E_{CL}\), \\
    Predefined linear guidance schedule \(\mathcal{G} = \{\lambda_1, \lambda_2, \ldots, \lambda_e, \ldots, \lambda_{E_{CL}}\}\)
    }
    
    \KwOutput{Trained model \(f_\theta\)}
    \kwInitialize{Pretrained model \(f_\theta\)}

    \For(){\(e \leq E_{CL}\)}{
        \(\lambda_e = \mathcal{G}(e)\) \\
        Extract \(\mathcal{S}_{\lambda_e} = \{(x'^{(j)}, y^{(j)}, \lambda^{(j)}) \mid \lambda^{(j)} = \lambda_e\}\) \\
        Gather new training set \(\mathcal{D}_e = \mathcal{S}_{\lambda_e} \cup \mathcal{D}_{\text{nh}} \cup \mathcal{D}_{\text{h}}\) \\
        Finetune model \(f_\theta\) with \(\mathcal{D}_e\)
    }

    \For(){\(E_{CL} < e \leq E\)}{
        Gather new training set \(\mathcal{D}_e = \mathcal{D}_{\text{nh}} \cup \mathcal{D}_{\text{h}}\) \\
        Finetune model \(f_\theta\) with \(\mathcal{D}_e\)
    }

    \caption{Training with \textbf{non-adaptive} curriculum strategy}
    \label{alg:DisCLnonadaptive}
\end{algorithm}

%% file: Algorithm_tex/adaptive.tex
\begin{algorithm}
    \SetKwInput{KwInput}{Input}
    \SetKwInput{kwInitialize}{Initialize}
    \SetKwInput{KwOutput}{Output}

    \KwInput{
    Image guidance levels \(\Lambda = \{\lambda_i \mid \lambda_i \in [0, 1]\}\), \\
    Non-hard samples \(\mathcal{D}_{\text{nh}} = \{(x^{(j)}, y^{(j)}, \lambda^{(j)} = 1)\}_{j=1}^N\), \\
    Syn-to-real spectrum data \(\mathcal{S} = \{(x'^{(j)}, y^{(j)}, \lambda^{(j)}) \mid \lambda^{(j)} \in \Lambda\}_{j=1}^M\), \\
    Combined training data \(\mathcal{D}_{\text{all}} = \mathcal{D}_{\text{nh}} \cup \{(x'^{(j)}, y^{(j)}, \lambda^{(j)}) \mid \lambda^{(j)} = 1\}\), \\
    Guidance validation set \(\mathcal{V} = \{(x'^{(j)}, y^{(j)}, \lambda^{(j)}) \mid \lambda^{(j)} \in \Lambda\}_{j=1}^m\), \\
    Total training epochs \(E\), curriculum cutoff epoch \(E_{CL}\), size of real-random set \(|D|\)
    }
    \KwOutput{Trained model \(f_\theta\)}
    \kwInitialize{Pretrained model \(f_\theta\)}
    \tcc{Note: \(\mathcal{V}\) has no overlap with \(\mathcal{D}_{\text{all}}\)}

    \For(){\(e \leq E_{CL}\)}{
        Compute true-class probability \(p_{\text{bef}}\) of model \(f_\theta\) on \(\mathcal{V}\) \\
        Sample a random-real set \(D\) from \(\mathcal{D}_{\text{all}}\) \tcc{contains only real data} 
        
        Train model \(f_\theta\) with \(D\) \\
        Compute true-class probability \(p_{\text{aft}}\) of model \(f_\theta\) on \(\mathcal{V}\) \\
        
        \(\lambda_e \gets \arg\max_{\lambda_i \in \Lambda} \left(p_{\text{aft}}(\lambda_i) - p_{\text{bef}}(\lambda_i)\right)\) \\
        Extract \(\mathcal{S}_{\lambda_e} = \{(x'^{(j)}, y^{(j)}, \lambda^{(j)}) \mid \lambda^{(j)} = \lambda_e\}\) \\
        Form training set \(\mathcal{D}_e = \mathcal{S}_{\lambda_e} \cup \mathcal{D}_{\text{nh}}\) \\
        Train model \(f_\theta\) with \(\mathcal{D}_e\)
    }
    
    \For(){\(E_{CL} < e \leq E\)}{
        Train model \(f_\theta\) with \(\mathcal{D}_{\text{all}}\)
    }

    \caption{Training with \textbf{adaptive} curriculum strategy}
    \label{alg:DisCLadaptive}
\end{algorithm}

%% file: Table_tex/hyperparameters.tex
\begin{table*}[ht]
\centering
\begin{tabular}{@{}c|l|r}
\toprule
& \textbf{Hyperparameter Name}     & \textbf{Value}  \\
\midrule
\parbox[t]{2mm}{\multirow{8}{*}{\rotatebox[origin=c]{90}{Synthetic Generation}}} & Text Guidance Scale \(w\)   & 10    \\
 & Noise Scheduler & DDIM \\
 & Stable Diffusion Denoising Steps       & 1000   \\
 & Stable Diffusion Checkpoint  & stabilityai/stable-diffusion-xl-refiner-1.0 \\
 & CLIP Filter Model & openai/clip-vit-base-patch32 \\
 & Filtering Threshold for iWildCam & 0.25 \\
 & Filtering Threshold for ImageNet-LT & 0.30 \\
 & GPU Used & Nvidia rtx5000 with 24GB \\
\midrule
\parbox[t]{2mm}{\multirow{9}{*}{\rotatebox[origin=c]{90}{ImageNet-LT}}} & Level of Image Guidances \(\lambda\) & \(\{0, 0.1, 0.3, 0.5, 1.0\}\) \\
 & CLIP Filtering Threshold  &  0.3\\
 & Batch Size for ResNet-10  &  128 \\
 & Learning Rate  &  1\(e\)-3 \\
 & Optimizer  & Adam \\
 & Scheduler  & Cosine \\
 & Training Epoch  & 65 \\
 & Training Epoch for Curriculum Learning & 60 \\
 & GPU Used & Nvidia rtx5000 with 24GB \\
\midrule
\parbox[t]{2mm}{\multirow{13}{*}{\rotatebox[origin=c]{90}{iWildCam}}} & Level of Image Guidances \(\lambda\) & \(\{0.5, 0.7, 0.9, 1.0\}\) \\
 & CLIP Filtering Threshold  &  0.25\\
 & Size of Dataset \(D\) & 30000 \\
 & Size of Guidance Validate Dataset \(S\) & 2000 \\
 & Batch Size for CLIP ViT-B/16  &  256 \\
 & Batch Size for CLIP ViT-L/16  &  200 \\
 & Learning Rate  & 1\(e\)-5 \\
 & Optimizer  & AdamW \\
 & Scheduler  & Cosine with Warmup \\
 & Warmup Step  & 500 \\
 & Training Epoch  & 20 \\
 & Training Epoch for Curriculum Learning & 15 \\
 & GPU Used & 2 Nvidia A100 with 80GB \\
\bottomrule
\end{tabular}
\vspace*{-0.2cm}

\caption{Hyperparameters and their values}
\vspace*{0.2cm}
\label{tab:hyperparameter}
\vspace*{-1\baselineskip}
\end{table*}

%% file: Figure_tex/ablation_guid.tex
\begin{figure*}[ht]
    \centering
    \subfloat{
        \includegraphics[width=0.45\textwidth]{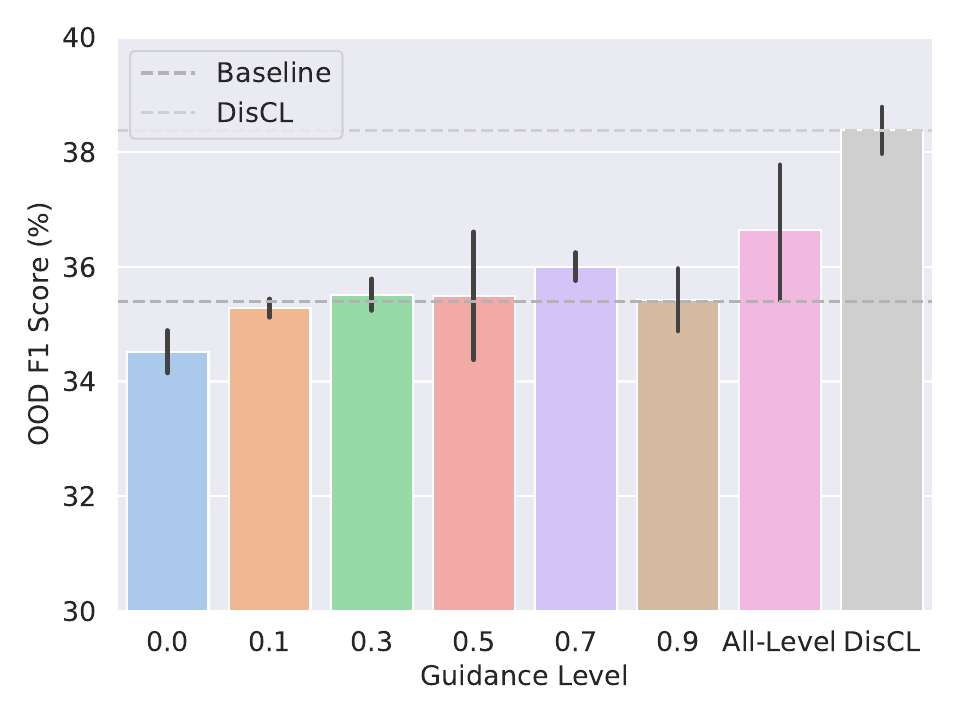}
    }
    \subfloat{
        \includegraphics[width=0.45\textwidth]{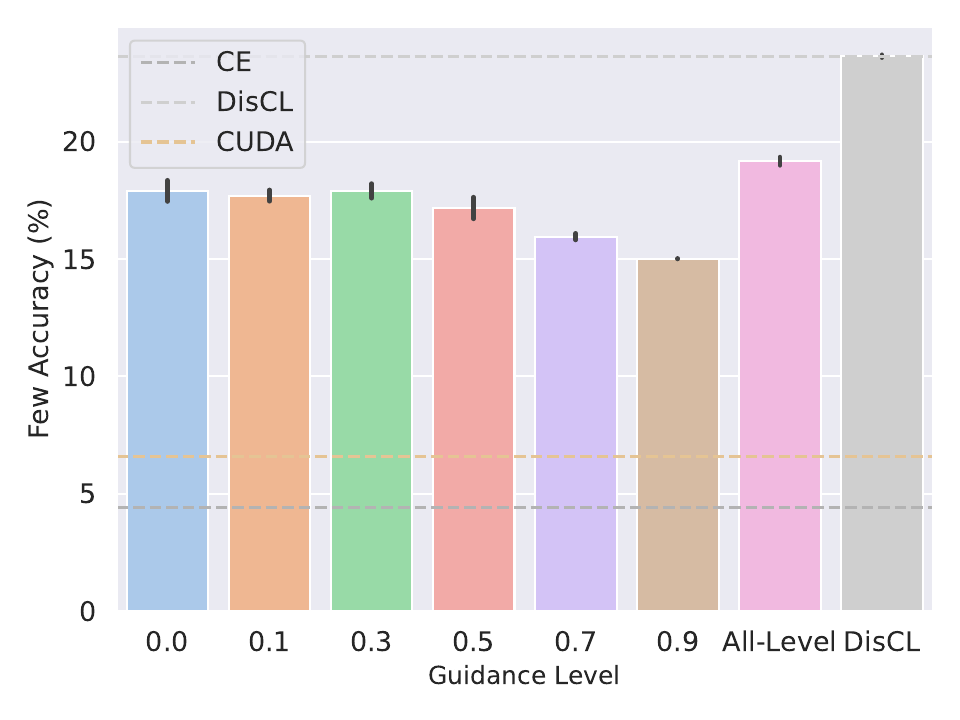}
    }
    \caption{Effect of Image Guidance (mixing syn+real). All-level experiments use the synthesis samples from all guidance scales selected for each task. \(0.5\) refers to only using synthetic data with guidance level \(\lambda=0.5\) for fine-tuning. Left: results on iWildCam. Right: results on ImageNet-LT}
    \label{fig:GuidanceDifference}
    \vspace*{-1\baselineskip}
\end{figure*}

%% file: Table_tex/compare_bs.tex
\begin{table}[h]
\vspace{-10pt}
    \centering
    \resizebox{0.85\columnwidth}{!}{%
    \begin{tabular}{c|ccccc}
        \toprule
        \(k\) & 10 & 50 & 100 & 150 & 200 \\
        \midrule
        $\text{Acc}_{\text{BS+DisCL}} - \text{Acc}_{\text{BS}}$ & 7.6\% & 6.0\% & 5.7\% & 5.2\% & 4.5\%\\
        \bottomrule
    \end{tabular}
    }
    \vspace{-7pt}
    \caption{\small{\textbf{Improvement in Accuracy on Worst-\(k\) classes in INLT}.}}
    \vspace{-10pt}
    \label{tab:worst-k}
\end{table}